\documentclass{adobe_research}
\usepackage[T1]{fontenc}
\usepackage{lmodern}
\usepackage{multirow}
\usepackage{soul}
\usepackage{pifont}
\usepackage[T1]{fontenc}
\usepackage[table]{xcolor}
\usepackage{graphicx}
\usepackage{amsmath,amssymb}
\usepackage{bbm}
\usepackage{makecell}
\usepackage[dvipsnames]{xcolor}
\usepackage{color, colortbl}
\usepackage{caption}
\usepackage{algorithm}
\usepackage{algpseudocode}
\usepackage{booktabs}
\usepackage{tabularx}
\usepackage{array}
\usepackage{ragged2e} 

\usepackage{xspace}
\usepackage{booktabs}
\makeatletter
\DeclareRobustCommand\onedot{\futurelet\@let@token\@onedot}
\def\@onedot{\ifx\@let@token.\else.\null\fi\xspace}

\makeatother

\usepackage{etoolbox}
\makeatletter
\patchcmd{\mymaketitle}{%
  \vskip 0.45cm
}{%
  \vskip 0.05cm
\noindent\begin{minipage}{\textwidth}
  \centering
  \includegraphics[width=1\textwidth,trim=0 18pt 0 18pt,clip]{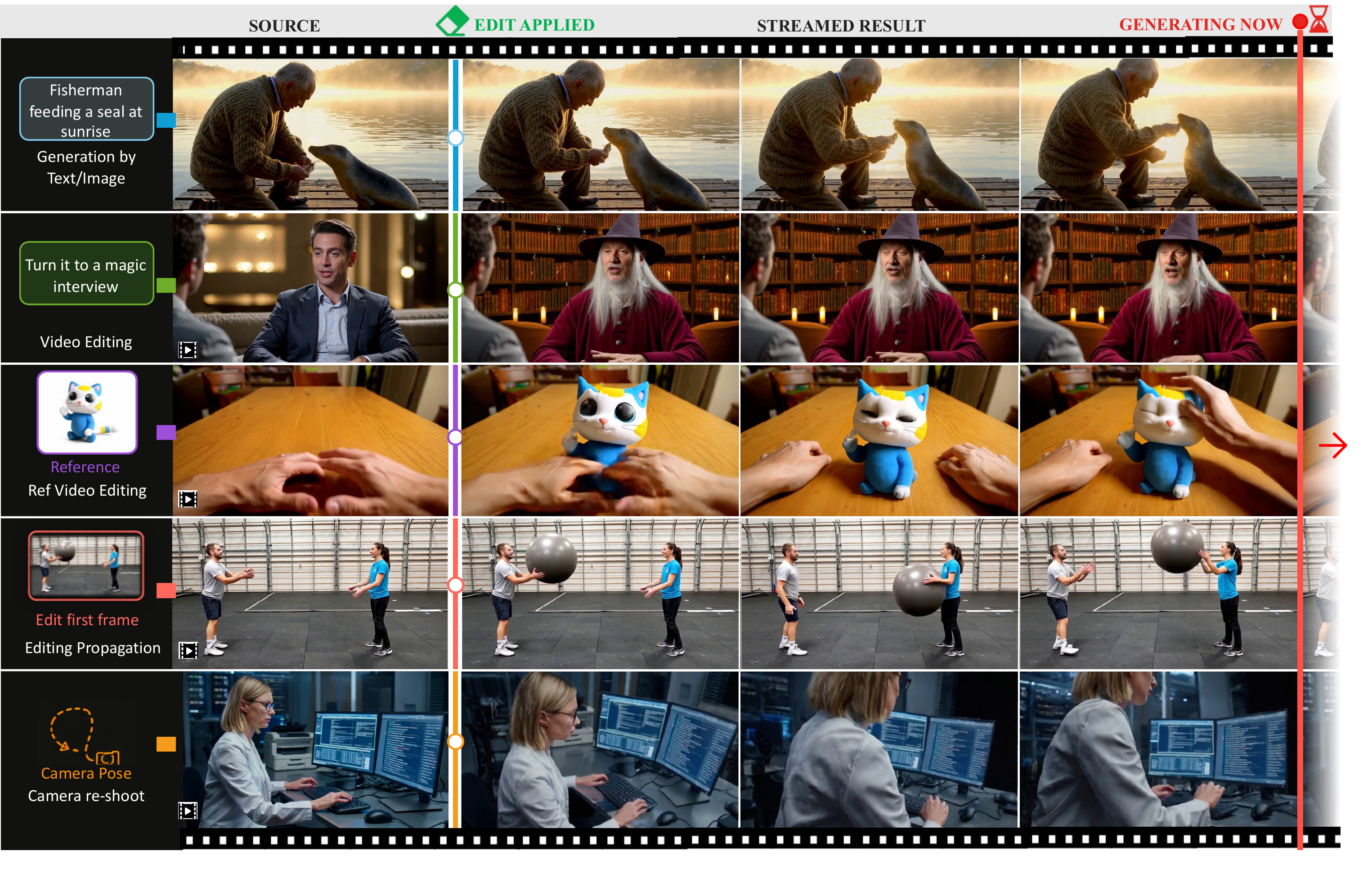}
  {\setlength{\abovecaptionskip}{-4pt}\setlength{\belowcaptionskip}{-2pt}%
   \captionsetup{type=figure,font=footnotesize}
   \captionof{figure}{\textbf{Interactive video generation and editing with EditStream}. Starting from text, images, source videos, or live camera, users can generate new content, apply edits, introduce a subject from a reference image, propagate a first-frame edit, or re-shoot a scene with explicit camera control. EditStream progressively produces the edited video through few-step autoregressive inference.}
  }
  \label{fig:teaser}
\end{minipage}
  \vskip 0.03cm
}{}{}

\patchcmd{\mymaketitle}{\tcbset{top=0.5cm}}{\tcbset{top=0.22cm}}{}{}
\patchcmd{\mymaketitle}{\tcbset{bottom=0.5cm}}{\tcbset{bottom=0.22cm}}{}{}
\patchcmd{\mymaketitle}{\vskip 0.5cm}{\vskip 0.12cm}{}{}
\makeatother

\usepackage{newtxtext}

\newcommand{\R}{\mathbb{R}}
\newcommand{\E}{\mathbb{E}}
\newcommand{\concat}{\mathrm{Concat}}
\newcommand{\stopgrad}{\mathrm{sg}}

\usepackage{algpseudocode}
\definecolor{catgray}{gray}{0.92}

\newcommand{\groupheader}[1]{%
  \rowcolor{gray!15}
  \multicolumn{7}{l}{\small\bfseries #1} \\
}
\newcommand{\tablegroupheader}[2]{%
\rowcolor{gray!15}
    \multicolumn{#1}{l}{\textbf{#2}} \\
}
\newcommand{\refviegroupheader}[1]{%
  \rowcolor{gray!15}
  \multicolumn{10}{l}{\hspace{3pt}\small\bfseries #1} \\
}

\newcommand{\vbenchgroupheader}[1]{%
  \rowcolor{gray!15}
  \multicolumn{12}{l}{\hspace{3pt}\small\bfseries #1} \\
}

\usepackage[dvipsnames]{xcolor} 
\definecolor{FutureOrange}{HTML}{EC866D}

\title{EditStream: A Unified Autoregressive Framework for Interactive Video Generation and Editing}

\author[1,*,\dagger]{Yuqian Zhou}
\author[1,2,*,\ddagger]{Zhenghong Zhou}
\author[1]{Zongze Wu}
\author[1]{Cameron Smith}
\author[1]{\\Richard Zhang}
\author[2]{Jiebo Luo}
\author[1]{Eli Shechtman}
\author[1]{Zhe Lin}

\affiliation[1]{Adobe Research}
\affiliation[2]{University of Rochester}

\contribution[*]{First authors in random order}
\contribution[\ddagger]{Work done during an internship at Adobe Research}
\contribution[\dagger]{Project lead}

\abstract{
\begin{center}
\textbf{Abstract}
\end{center}
\vspace{0.5\baselineskip}

\begingroup
\justifying
Interactive video generation and editing are becoming increasingly important for creative design. In this report, we introduce EditStream—a unified framework for interactive video generation and editing. EditStream unifies multiple video creation and manipulation tasks within a single DiT-based model through flexible task-specific conditioning, and further transforms it into a fast, few-step autoregressive model for efficient streaming. It supports Text-to-Video, Image-to-Video, Video-to-Video, Editing Propagation, Reference-guided Video Editing, and Camera Pose Change, enabling flexible control over video generation, transformation, and editing within one system. To make the unified model practical for interactive use, we develop a two-stage distillation approach that combines Velocity Moment Matching (VMM) with autoregressive unrolling. VMM matches conditional velocity moments at student-reached intermediate states to preserve generation quality and motion, while unrolling exposes the student to its own autoregressive predictions to improve temporal stability. Together, they alleviate common challenges in few-step autoregressive video generation, including over-saturation, degraded motion, temporal instability, and complex training. EditStream provides a practical and scalable solution that bridges high-quality diffusion-based video models with interactive creative workflows.
\par

\vspace{0pt}
{\scriptsize\raggedright\textbf{Project page:} \href{https://real-time-video-research.github.io/editstream/}{\texttt{real-time-video-research.github.io/editstream/}}\par}

\endgroup
}

\begin{document}

\maketitle

\section{Introduction}
\label{sec:introduction}

Recent works on video diffusion models have improved the fidelity, diversity, and semantic quality of video generation. They enable realistic text-to-video and image-to-video generation at high resolutions and with long contexts \citep{polyak2024movie,wan2025,hacohen2026ltx,longlive,yuan2026helios}. However, creative workflows demand more than generating a complete video from text prompts. A user may want to begin with text, animate an image, manipulate an existing video, propagate an edited image from one frame to the entire video, insert a reference subject into the video, or redirect the camera trajectory. These operations are also better performed in an interactive way: the model should respond as the video is generated or played. This shift from an \emph{offline video model} to an \emph{interactive streaming video model} introduces a major challenge: a useful system must support multiple controls, preserve high visual and temporal quality, and operate in a causal streaming manner at the same time.

Existing research has made substantial progress along each of these dimensions, but largely in isolation. Some works unify video generation and editing within a single DiT \cite{dit}. EditVerse \citep{ju2025editverse} concatenates text, image, and video tokens to formulate in-context learning for video and image generation and editing. UniVideo \citep{wei2025univideo} combines multimodal understanding with DiT to interpret complex and compositional instructions. VACE \citep{jiang2025vace} also proposes a unified multimodal generation model. These works claim that training data plays a vital role in quality improvement. The curated data includes synthetic pairs \cite{cheng2024consistent,zi2026senorita} and region-based pairs \cite{zhang2025region}. Nevertheless, most unified models remain designed for offline denoising based on bidirectional attention. Moreover, simply treating every visual condition as additional context tokens can substantially increase the sequence length and attention cost, particularly when the source video, spatial mask, geometric render, and target video are all temporally and spatially aligned.

Meanwhile, autoregressive (AR) video diffusion has become a promising route toward streaming generation. Diffusion Forcing \citep{chen2024diffusionforcing} first combines next-token prediction with full-sequence diffusion training and enables different temporal blocks to be generated at different noise levels. CausVid is proposed to convert a bidirectional video diffusion model into a few-step causal student using Diffusion Forcing initialization and Distribution Matching Distillation (DMD) \citep{yin2025slow}. After that, Self-Forcing further reduces the train-inference gap by unrolling the student’s own generated history sequence \citep{huang2026self}. It has been widely adopted in streaming video editing systems such as EgoEdit, LiveEdit, and SANA-Streaming \citep{li2026egoedit,wang2026liveedit,zhao2026sana}, as well as world models \citep{hong2025relic}, and has been generalized to long video generation by addressing long-horizon drifting issues \citep{li2026rolling,cui2025self}.

However, it remains challenging to convert a strong bidirectional teacher diffusion model into a few-step causal student. Previous works mostly involve multiple stages, including offline ODE data collection \cite{huang2026self,yin2025slow}, ODE initialization \cite{huang2026self,yin2025slow,zhu2026causalforcing}, teacher-to-causal conversion \cite{zhao2026causalforcingp,zhu2026causalforcing,zheng2026causalrcm}, consistency distillation as initialization \cite{zheng2026causalrcm,zhao2026causalforcingp}, teacher forcing with artifact augmentation \cite{yuan2026helios}, and self-forcing unrolling \cite{huang2026self,zheng2026causalrcm,zhu2026causalforcing,zhao2026causalforcingp}. These approaches are effective, but can be difficult to reproduce, especially when the model unifies multiple different video tasks altogether. Also, regarding the distillation objectives, DMD \cite{yin2024onestep,yin2024improved} tends to suffer from mode collapse despite generating sharp results through distribution matching, while consistency distillation \cite{zhao2026causalforcingp} or flow-map methods \cite{geng2025meanflow} directly regress teacher endpoints or transitions but do not correct the states reached by the student sampler. Simplifying this pipeline with better objectives therefore becomes essential.

In this work, we present \textbf{EditStream}, a unified autoregressive framework for interactive video generation and editing. EditStream formulates diverse video tasks, including Text-to-Video (\textbf{T2V}), Image-to-Video (\textbf{I2V}), Video-to-Video (\textbf{V2V}) editing, first-frame editing propagation (\textbf{EditProp}), reference-guided Video-to-Video editing (\textbf{refV2V}), and camera pose control (\textbf{ReShoot}) within one diffusion-transformer backbone. We unify them using the proposed dual-type conditioning interface. It uses channel concatenation for pixel-aligned conditions and token concatenation for non-aligned reference images and edited frames. To make the unified model streamable, we introduce a practical bidirectional-to-autoregressive conversion recipe based on \textbf{Velocity Moment Matching (VMM)}. The recipe follows warm-up and unrolling phases to stabilize the semantics while reducing the exposure gap without offline ODE-pair preparation or teacher causalization. Energy Annealing (\textbf{EA}) is proposed to mitigate over-saturation caused by teacher classifier-free guidance during distillation. Experiments demonstrate that the causal student preserves most of the unified teacher's capabilities while enabling few-step autoregressive inference with the proposed simple yet efficient recipe. Our main contributions towards interactive streaming video generation and editing are summarized as follows:
\begin{itemize}
\item
\textbf{A unified formulation for efficient teacher:} We formulate T2V, I2V, V2V, RefV2V, EditProp, and ReShoot tasks within a single DiT using a dual-type conditioning interface. This unified formulation enables different tasks to benefit from mixed training.

\item
\textbf{A simple bidirectional-to-autoregressive conversion recipe.}
We introduce Velocity Moment Matching (\textbf{VMM}) distillation algorithm, building upon DMD and Moment Matching (MM). We modify the timestep sampling strategy and training objective to enable faster convergence by correcting student transitions with the teacher–auxiliary velocity residual evaluated at student-induced states. We preserve a high-quality full-attention teacher and convert it into a causal student using a two-stage recipe. Combined with Energy Annealing (EA), this recipe improves motion dynamics, temporal stability, and color quality during autoregressive generation.

\end{itemize}
\section{Related Works}

\textbf{Unified Video Generation and Editing.}
Recent advances in image, video, and multimodal generation have increasingly shifted toward generalist models, with the goal of enabling a single model to perform diverse tasks such as generation and editing. To unify these heterogeneous tasks, researchers have explored unified modeling paradigms that extend from images ~\cite{geyer2024tokenflow,deng2025emerging,chen2025unireal} to videos. In the video domain, representative works such as VACE~\citep{jiang2025vace} and EditVerse~\citep{ju2025editverse} unify image and video generation/editing within a single framework by designing a shared conditional interface. Existing approaches typically incorporate conditioning signals through mechanisms such as cross-attention~\cite{yang2026omni}, ControlNet~\citep{zhang2023adding,zhou2026tri,liu2025generative}, channel concatenation~\citep{chen2026skyreels,wu2026loomvideo}, or token concatenation~\cite{ju2025editverse, liu2026tide, cai2026omnivcus}, allowing different tasks to be represented in a unified manner. Meanwhile, researchers leverage the strong multimodal understanding capabilities of multimodal large language models (MLLMs) to bridge generation and understanding ~\citep{team2025kling,chen2026vino,yang2026omni,team2026bernini,wei2025univideo}. By jointly modeling content generation and multimodal comprehension, these methods provide more accurate semantic guidance for generation and editing, thereby improving the fidelity, controllability, and overall quality of multimedia content.

EditStream builds upon these unified conditioning paradigms while prioritizing efficient inference as a primary design objective. Since reducing the number of conditioning tokens is crucial for improving inference efficiency in diffusion transformers, channel concatenation is adopted as the default conditioning mechanism whenever possible. However, channel concatenation is inherently limited to pixel-aligned inputs and cannot effectively represent non-spatial conditions. To address this limitation, we propose a dual-type conditioning scheme that combines the strengths of both conditioning paradigms. Specifically, pixel-aligned inputs, such as source videos, are injected through channel concatenation to minimize token overhead, whereas non-pixel-aligned conditions, such as reference images, are represented as additional context tokens via token concatenation. This hybrid design enables EditStream to efficiently unify a wide range of video generation and editing tasks while preserving the flexibility to incorporate heterogeneous conditioning signals.

\textbf{Streaming Autoregressive Video Generation and Editing.} High-quality video generation models typically denoise all frames within a fixed-length clip jointly. This formulation allows each frame to exploit the full spatiotemporal context of the clip, enabling globally coordinated refinement and improved temporal consistency. However, its computational and memory costs grow rapidly with video duration and spatial resolution, making it unsuitable for low-latency or long-form video generation. Early studies, including Rolling Diffusion~\cite{ruhe2024rollingdiffusion} and FIFO~\citep{kim2024fifo}, addressed this limitation by progressively generating video sequences through rolling sliding windows or denoising queues. Nevertheless, frames within the same window may still interact bidirectionally, as both historical and future frames participate in the denoising process. A strictly causal formulation instead requires the current frame or chunk to attend only to previously generated frames or chunks. Such a design makes it possible to cache and repeatedly reuse the key–value representations of the historical context, substantially reducing redundant computation during streaming inference.

Diffusion Forcing~\citep{chen2024diffusionforcing} unified full-sequence denoising and frame-wise progressive denoising within a single training framework by assigning a different noise level to each frame. Teacher forcing~\cite{williams1989learning,huang2026self,zhu2026causalforcing} can simplify the conditioning process by replacing the historical context with clean tokens and applying noise only to the current frame or chunk. Building on these ideas, CausVid~\citep{yin2025slow} demonstrated how a bidirectional video diffusion model can be converted into a strictly causal streaming model. Specifically, it combines an initialization procedure based on ordinary differential equation trajectories and Diffusion Forcing with distribution-matching distillation (DMD)~\citep{yin2024onestep,yin2024improved}, thereby establishing an effective pathway for converting pretrained video models into models capable of low-latency streaming inference.

A subsequent line of work has focused on reducing the exposure bias of streaming video models by better aligning their training and inference distributions. During training, a model is typically conditioned on clean ground-truth context, whereas at inference time, the student model must autoregressively condition on its own generated frames. Any distributional discrepancy between generated frames and clean ground truth—such as degraded visual quality, abrupt color changes, or accumulated motion errors—can therefore propagate through the sequence and cause progressively more severe degradation. Self-Forcing~\citep{huang2026self} first addressed this discrepancy by unrolling sequences generated by the student itself and applying DMD-based distillation to the resulting trajectories. Causal Forcing~\citep{zhu2026causalforcing} subsequently highlighted the importance of teacher causality within this framework. Its follow-up, Causal Forcing++~\citep{zhao2026causalforcingp}, improved the initialization strategy and achieved higher generation quality with fewer sampling steps. Causal-rCM~\citep{zheng2026causalrcm} reached a related conclusion and provided a theoretical analysis of how different initialization schemes affect the final generation performance. More recent works have also extended this self-rollout paradigm to one-step generation, like One-Forcing~\citep{feng2026one} and AAD-1~\citep{li2026aad}, further improving the efficiency of inference.

Many recent studies have extended these causal generation frameworks to long-form video creation. Their central challenges include the continuous accumulation of generation errors, the compression and retrieval of long-context memory, and the generalization of models trained on short clips to substantially longer inference horizons. Methods such as Rolling Forcing~\citep{liu2025rollingforcing}, Context Forcing~\citep{chen2026context}, Long Live~\citep{longlive,longlive_2.0,longlive_rag}, Self-Forcing++~\cite{cui2025self}, TetherCache~\citep{meng2026tethercache}, DySink~\citep{ye2026dysink} and Rolling Sink~\citep{li2026rolling} have explored complementary solutions involving attention sinks, rotary positional embedding design, memory representations, and cache-management strategies. These techniques improve the stability of long-horizon video generation and are largely complementary to EditStream. In particular, they could be incorporated into EditStream to support substantially longer video-editing sessions without introducing additional quality degradation.

Streaming video editing, which is most closely related to our setting, has also attracted increasing attention. Early systems commonly combined image diffusion models with additional temporal modeling modules to enable online editing~\citep{xing2024live2diff,liang2025looking,chen2025streaming}. More recently, the field has shifted toward constructing inherently causal video editors. Representative efforts include RFDM~\citep{salehi2026rfdm}, SANA-Streaming~\citep{zhao2026sana}, LiveEdit~\citep{wang2026liveedit}, EgoEdit~\citep{li2026egoedit}, VACE~\cite{jiang2025vace,fosdick2026adapting}, and commercial systems developed by companies such as Decart. These approaches investigate different strategies for unifying streaming editing tasks and distilling offline or bidirectional editing models into efficient causal models.

Building on these advances, EditStream fully leverages the diversity and scale of open-source data while complementing it with a carefully collected and filtered internal dataset. It unifies several important video generation and editing tasks within a single framework and investigates bidirectional-to-causal conversion across multiple task formulations. In contrast to prior work, EditStream emphasizes both broad multi-task unification and a simpler, more efficient, and more stable conversion-and-distillation recipe. Moreover, our approach is orthogonal to long-horizon memory and cache-management methods. These techniques can therefore be integrated with EditStream to further improve its scalability, temporal stability, and practical efficiency in real-world streaming video-editing deployments.

\section{Unified Video Generation and Editing}
\begin{figure*}[t]
    \centering
    \includegraphics[width=\textwidth]{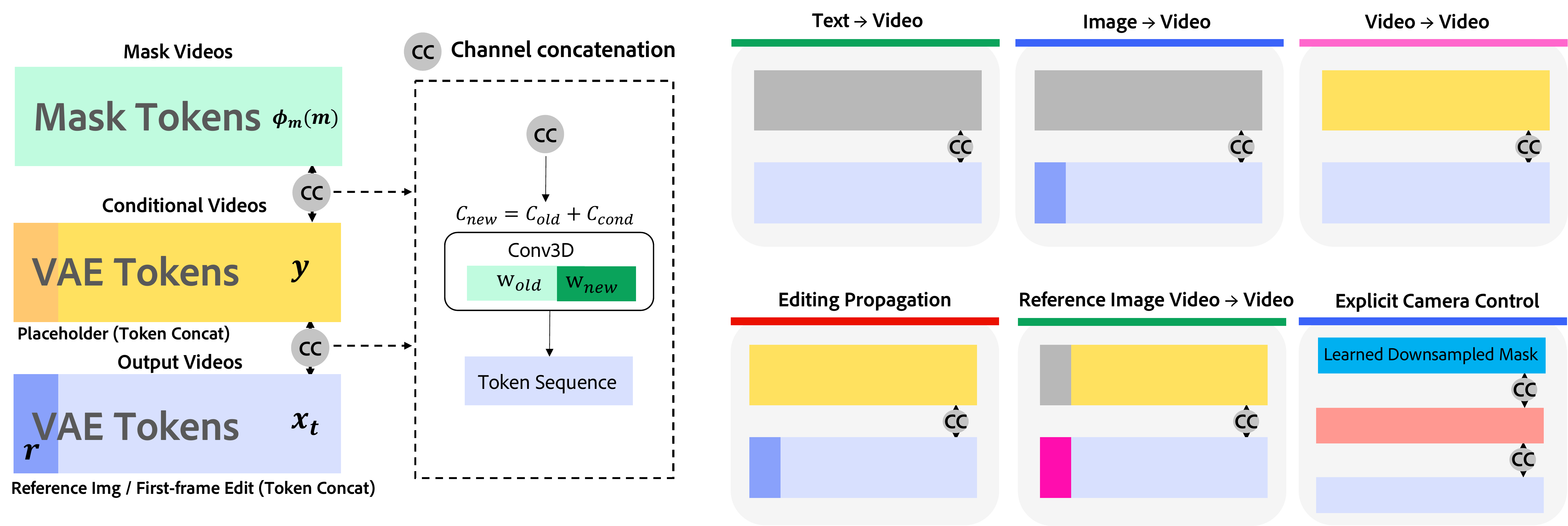}
    \caption{EditStream adopts a dual-type conditioning interface for unified video generation and editing: pixel-aligned conditions (e.g., source latent video or camera-warped inputs $\mathbf{y}$, and mask $\phi_m(\mathbf{m})$) are injected via compact channel concatenation $\tilde{\mathbf{x}}_t=\concat_{\mathrm{ch}}[\mathbf{x}_t,\mathbf{y},\phi_m(\mathbf{m})]$ to avoid increasing attention length, while non-aligned reference signals (e.g., a reference image or the first frame latent $\mathbf{r}$ for I2V/refV2V/EditProp) are provided by sequence token concatenation $\mathbf{x}^{*}_t=\concat_{\mathrm{tk}}[\mathbf{r},\mathbf{x}_t[:-1]]$ to better leverage self-attention for in-context learning.}
    \label{fig:taskconcat}
\end{figure*}

\subsection{Problem formulation}

We unify video generation and editing tasks in a latent video diffusion model. Let $\mathbf{x}_0 \in \R^{T \times C \times H \times W}$ denote the clean video target latent after the VAE, where $T$ is the number of latent frames, $C$ is the latent channel dimension, and $H,W$ are latent spatial dimensions. A diffusion or flow-matching training sample is produced by corrupting $\mathbf{x}_0$ with Gaussian noise to obtain $\mathbf{x}_t$:
\begin{equation}
\mathbf{x}_t = (1-\sigma_t)\mathbf{x}_0 + \sigma_t \boldsymbol{\epsilon}, \quad \boldsymbol{\epsilon}\sim\mathcal{N}(0,I), \quad \sigma_t\in[0,1],
\end{equation}
where $\sigma_t$ is the factor indicating noise levels. The model predicts a velocity field
\begin{equation}
\mathbf{v}_\theta(\mathbf{x}_t,t,\mathcal{C}) \approx \frac{d\mathbf{x}_t}{d\sigma_t} = \boldsymbol{\epsilon}-\mathbf{x}_0,
\end{equation}
conditioned on a task-dependent condition context and control signal set $\mathcal{C}$. The training objective can be written as:
\begin{equation}
\mathcal{L}_{\mathrm{fm}} =
\E_{\mathbf{x}_0, t, \boldsymbol{\epsilon}, \mathcal{C}}
\left[
\left||
\mathbf{v}_\theta(\mathbf{x}_t, t, \mathcal{C})
-
\mathbf{v}^{*}(\mathbf{x}_t, t, \mathbf{x}_0)
\right||_2^2
\right],
\end{equation}
where $\mathbf{v}^{*}$ is the target velocity defined by the underlying flow-matching parameterization.

Different generation and editing tasks share the same modeling and differ only in the content of $\mathcal{C}$. Therefore, a unified model can be trained by mixing the data from different tasks through a commonly shared conditioning interface. To avoid ambiguity, special task tokens are needed for some tasks. For example, editing propagation may share the same first-frame input as reference-guided Video-to-Video (refV2V), while requiring different model behaviors, so special task tokens will be helpful to differentiate the task. Camera control is modeled using a warped point cloud as the input source video. Because zero-valued pixels may represent either missing reprojected geometry or genuinely dark observed content, we additionally provide a binary validity mask and a task token. In EditStream, task tokens are input in the form of a special text prompt prefix.

Table~\ref{tab:task_formulation} summarizes the EditStream task space. The model optionally receives the text prompt, source video, first frame for propagation, mask, or single reference image and predicts a target video according to the given conditions and special task tokens. This formulation supports both generation and editing. T2V synthesizes new content, whereas I2V, V2V, refV2V, EditProp and ReShoot preserve or transform existing contents. The same model is expected to learn when to preserve, when to modify, and how to balance the input controls without ambiguity.

\begin{table}[htbp]
\centering
\caption{Unified task formulation in EditStream. All tasks share the same video diffusion backbone and generate a video; they differ only in their conditional inputs and special task tokens.}
\label{tab:task_formulation}
\resizebox{\textwidth}{!}{
\begin{tabular}{llll}
\toprule
Task & Abbrev. & Conditional Inputs & Special Task Tokens\\
\midrule
Text-to-Video & T2V & Text prompt & --\\
Image-to-Video & I2V & First-frame image; (optional) Text prompt & --\\
(Egocentric) Video-to-Video & (Ego)V2V & Source (general/egocentric) video; Text prompt & --\\
Editing Propagation & EditProp & Source video; First-frame edit; (optional) Text prompt & [Editing Propagation]\\
Reference-guided Video-to-Video & refV2V & Source video; Reference image; Text prompt & [Reference Guidance]\\
Explicit Camera Pose Control & ReShoot & Re-projected source video; Mask; (optional) Text prompt & [Camera Control]\\
\bottomrule
\end{tabular}
}
\end{table}

\subsection{Dual-type Conditional Modeling}
EditStream introduces a dual-type of conditioning scheme for different video generation and editing tasks, combining compact channel concatenation and sequence token concatenation. 

For commonly-used streaming video editing cases, the input, control signals, and the output videos are pixel-aligned. For example, suppose that the input is sourced from a user's web camera. The user will expect the output as a mirror to faithfully reflect the motion and action to achieve better interactive experience. In this case, the target latent video usually has the same temporal and spatial grid as the source latent video and the control signal. For tasks like explicit camera control, since we use the warped pixels according to the target camera pose, the pixel-aligned modeling still applies. For such controls, adding them as extra sequence tokens is inefficient because it may dramatically increase the attention length as the conditions grow. Therefore, EditStream uses channel concatenation for all pixel-aligned conditions as shown in Figure \ref{fig:taskconcat}, illustrating the compact conditions for different tasks:

\begin{equation}
\tilde{\mathbf{x}}_t =
\concat_{\mathrm{ch}}
\left[
\mathbf{x}_t,
\mathbf{y},
\phi_m(\mathbf{m})
\right],
\end{equation}
where $\mathbf{y}$ is a pixel-aligned conditional video latent representing the source videos to be edited, $\mathbf{m}$ is a mask video in original resolution, and $\phi_m$ is a lightweight learnable downsampler that maps the mask video to the latent resolution. The patch-embedding layer is expanded to accept the additional channels. If the original patch embedding has weights
\begin{equation}
W_{\mathrm{old}} \in \R^{C_{\mathrm{out}} \times C_{\mathrm{in}} \times k_t \times k_h \times k_w},
\end{equation}
then the expanded patch embedding is initialized as
\begin{equation}
W_{\mathrm{new}} =
\left[
W_{\mathrm{old}};
\mathbf{0}
\right] \in \R^{C_{\mathrm{out}} \times (C_{\mathrm{in}} + C_{\mathrm{cond}}) \times k_t \times k_h \times k_w},
\end{equation}
where the original channels are copied and the newly added conditional channels are zero-initialized. This preserves compatibility with the pretrained text-to-video backbone and allows the model to quickly learn how to use visual controls. We also zero out the additional condition channels for tasks which do not have conditions, like Text-to-Video. 

However, for tasks like image-to-video or reference image guided video-to-video or editing propagation from single image, we instead use token concatenation to better leverage the self-attention for in-context learning. Specifically,

\begin{equation}
\mathbf{x}^{*}_t =
\concat_{\mathrm{tk}}
\left[\mathbf{r},
\mathbf{x}_t[:-1]
\right],
\end{equation}

where $\mathbf{r}$ is the VAE latent of the reference image or the first frame.

\section{Velocity Moment Matching (VMM)}
EditStream introduces a distillation algorithm named Velocity Moment Matching (VMM). Moment Matching Distillation (MM)~\cite{salimans2024multistep}, like DMD~\cite{yin2024onestep}, is a distribution-matching distillation method. Unlike DMD, which independently re-noises a complete student output to an arbitrary noise level and matches the resulting marginal distributions, MM first lets the student perform a coarse denoising transition and then matches conditional moments at the resulting noisy state. Specifically, starting from a noisy state ($x_t$), the student first predicts a clean-data estimate ($\tilde{x}$) with one forward pass and then performs a coarse transition to a less noisy state ($x_s$), where ($s<t$). Given $x_s$, an auxiliary denoiser estimates the conditional mean ($\mathbb{E}{g}[\tilde{x}\mid x_s]$) under the student-induced distribution $g$, while the pretrained teacher estimates the corresponding conditional mean ($\mathbb{E}{q}[x\mid x_s]$) under the target distribution $q$. MM optimizes the student by matching these two conditional means, encouraging each coarse transition to preserve the target distribution.

VMM builds upon this MM principle by formulating moment matching directly in the model’s velocity-output space. This design aligns the distillation objective with the native parameterization of flow-matching models. Although matching conditional velocity moments is theoretically equivalent to matching clean-sample moments under an invertible linear parameterization, the (x)-space formulation introduces timestep-dependent gradient scaling through both $t$ and $s$. Direct velocity matching removes this implicit scaling across time steps. Moreover, because velocity directly specifies the direction in which a noisy state evolves, optimizing in velocity space provides a more direct learning signal for each coarse transition. Together with redesigned timestep sampling and training objectives, these properties simplify the formulation and contribute to faster convergence for flow-matching-based video models. More details and the design of the scheme are explained below, and in the Appendix.

\subsection{Distillation Scheme}
\textbf{Model Roles.} We use three model roles to formulate VMM distillation, the same as moment matching distillation. 
\begin{itemize}
\item \textbf{Teacher} \(T\): a frozen video diffusion model. It keeps its original full attention and serves as a target.
\item \textbf{Auxiliary model} \(A\): a trainable model. It estimates the velocity field on the student's distribution and stabilizes moment-matching targets.
\item \textbf{Student} \(S\): the target distilled video model. For bidirectional inference, the student shares the same architecture as the teacher. For streaming inference, it uses causal block attention instead.
\end{itemize}

Suppose we denote the velocity predictions of the student, teacher, and auxiliary model as \(v_S\), \(v_T\), and \(v_A\)
 respectively. Given a noisy latent
\(\mathbf{x}_t\) at timestep $t$, the student first predicts
\[
v_t^S = v_S(\mathbf{x}_t, t, \mathcal{C}),
\]
and performs an ODE update from timestep \(t\) to a lower timestep \(s\):
\begin{equation}
\mathbf{x}_s
=
\mathbf{x}_t
+
(\sigma_s - \sigma_t) v_t^S .
\end{equation}
For the student loss, we use the student velocity induced by $x_s$,
denoted by \(\bar v_S^{t\to s}\). In the deterministic ODE case,
\(\bar v_S^{t\to s}=v_t^S\). The teacher and auxiliary model are then evaluated
at the same fake student state \((\mathbf{x}_s,s)\).

The auxiliary model is trained to track the detached student velocity on the
student's current fake distribution:
\begin{equation}
\mathcal{L}_{A}
=
\E_{t,s}
\left[
\left\|
\left(
v_A(\mathbf{x}_s, s, \mathcal{C})
-
\stopgrad(\bar v_S^{t\to s})
\right)
\right\|_2^2
\right],
\end{equation}

,where $sg$ means stop gradients, $\mathcal{C}$ is the conditional context. The student is trained with the velocity moment-matching target,
\begin{equation}
\tilde{v}
=
\stopgrad\left(
\bar v_S^{t\to s}
+
v_T(\mathbf{x}_s, s, \mathcal{C})
-
v_A(\mathbf{x}_s, s, \mathcal{C})
\right),
\end{equation}
so the student objective is
\begin{equation}
\mathcal{L}_{S}
=
\E_{t,s}
\left[
\left\|
\left(
\bar v_S^{t\to s}
-
\tilde{v}
\right)
\right\|_2^2
\right].
\end{equation}
Here the teacher provides the target velocity field, while the auxiliary model
estimates the velocity field of the student's fake distribution. Thus, the
correction term \(v_T-v_A\) plays the role of a real-fake score difference, but
is expressed as a velocity residual along the student's ODE trajectory.

For few-step distillation, we also modify timestep sampling to match the
student inference-time schedule. Instead of uniformly sampling \(t\) and \(s\) from the entire timestep range, we sample \(t\) from the discrete student schedule
\(\{\tau_1,\tau_2,\tau_3,\tau_4\}\), and then sample \(s\) inside the segment
between \(\tau_i\) and the next inference timestep. The proposed sampling method improves convergence efficiency compared to the MM \cite{salimans2024multistep}, where \(t\) is sampled randomly, and \(s\) is sampled within a margin smaller than \(t\). The detailed algorithm is shown in Algorithm \ref{alg:vmm}. More analysis of the distillation objectives and comparison with other methods can be found in the Appendix. 
\begin{figure}[t]
\centering
\begin{minipage}{0.88\linewidth}

\begin{algorithm}[H]
\caption{Velocity Moment Matching (VMM)}
\label{alg:vmm}
\small

\begin{algorithmic}[1]
\Require Frozen teacher $T$; trainable auxiliary $A$ and student $S$
\Require Context $\mathcal{C}$; schedule
         $\{\tau_i\}_{i=0}^{J}$, $\tau_i > \tau_{i+1}$

\For{each training iteration}
    \State Sample $i \sim \mathrm{Unif}\{0,\ldots,J-1\}$
    \State Set $t \gets \tau_i$
    \State Sample $s \sim \mathcal{U}(\tau_{i+1},\tau_i)$
    \State Sample noisy latent $\mathbf{x}_t$ from ground truth data sample

    \State $v_t^S \gets v_S(\mathbf{x}_t,t,\mathcal{C})$
    \State $\mathbf{x}_s
        \gets \mathbf{x}_t + (\sigma_s-\sigma_t)v_t^S$
    \State $\bar{v} \gets v_t^S$

    \State $v_T^s \gets v_T(\mathbf{x}_s,s,\mathcal{C})$
    \State $v_A^s \gets v_A(\mathbf{x}_s,s,\mathcal{C})$

    \Statex \textbf{Auxiliary update}
    \State $\mathcal{L}_A
        \gets \left\|v_A^s-\operatorname{sg}(\bar{v})\right\|_2^2$
    \State $A \gets A-\eta_A\nabla_A\mathcal{L}_A$

    \Statex \textbf{Moment-matching target}
    \State $\tilde{v}
        \gets \operatorname{sg}\!\left(\bar{v}+v_T^s-v_A^s\right)$

    \Statex \textbf{Student update}
    \State $\mathcal{L}_S
        \gets \left\|\bar{v}-\tilde{v}\right\|_2^2$
    \State $S \gets S-\eta_S\nabla_S\mathcal{L}_S$
\EndFor
\end{algorithmic}

\end{algorithm}
\end{minipage}
\end{figure}

\subsection{Energy Annealing Strategies}
During teacher inference and distillation, we find that the distilled results may have over-saturation issues when the classifier-free guidance scale is set to a large value, and this issue propagates to the student. To mitigate this issue, we propose an energy annealing (EA) strategy, which benefits both teacher inference and distillation.

The teacher predicts velocity, so we first convert its velocity predictions into clean-sample $x_0$ predictions, $\hat{x}_0^{c} = x_t - \sigma_t v_\theta^{c}$ and $\hat{x}_0^{u} = x_t - \sigma_t v_\theta^{u}$. This is because color and saturation statistics are more meaningful in $x_0$-space. Let $\Delta x = \hat{x}_0^{c} - \hat{x}_0^{u}$ denote the guidance direction and $w$ the guidance scale, so that the standard CFG-guided prediction is formulated as $\hat{x}_0^{\mathrm{cfg}} = \hat{x}_0^{c} + (w-1)\Delta x$.

We need to find a reference point for $\hat{x}_0^{\mathrm{cfg}}$ to match in order to reduce saturation. We compute this reference prediction by applying a smaller guidance scale, $\hat{x}_0^{\mathrm{ref}} = \hat{x}_0^{c} + \eta_{\mathrm{EA}}(w-1)\Delta x$, where $0 < \eta_{\mathrm{EA}} < 1$. We then rescale the per-channel mean $\mu(\cdot)$ and standard deviation $\sigma(\cdot)$ of $\hat{x}_0^{\mathrm{cfg}}$---pooled over the frame and spatial axes so that the same correction is applied uniformly to every frame and pixel of a channel---to match those of $\hat{x}_0^{\mathrm{ref}}$:
\begin{equation}
\hat{x}_0^{\mathrm{EA}}
=
\frac{\sigma\bigl(\hat{x}_0^{\mathrm{ref}}\bigr)}
{\sigma\bigl(\hat{x}_0^{\mathrm{cfg}}\bigr) + \epsilon}
\Bigl(
\hat{x}_0^{\mathrm{cfg}}
-
\mu\bigl(\hat{x}_0^{\mathrm{cfg}}\bigr)
\Bigr)
+
\mu\bigl(\hat{x}_0^{\mathrm{ref}}\bigr).
\label{eq:ea-energy}
\end{equation}

\begin{equation}
v^{\mathrm{EA}} = \frac{x_t - \hat{x}_0^{\mathrm{EA}}}{\sigma_t}
\label{eq:ea-energy2}
\end{equation}

Now, $\hat{x}_0^{\mathrm{EA}}$ can be used to replace $\hat{x}_0^{\mathrm{cfg}}$ as the teacher output, which can both preserve the sharper structural details and textures provided by the full guidance scale $w$ and control its color and saturation to match those of the reference $\hat{x}_0^{\mathrm{ref}}$.

Moderate energy annealing ($\eta_{\mathrm{EA}} \approx 0.7$ in our experiments) improves saturation and color stability. However, a smaller factor may also over-suppress the overall guidance effect. A balanced factor needs to be manually tuned for different models.

\section{From Bidirectional Diffusion to Autoregressive Streaming}

Video models with bidirectional attention leverage full-sequence context to include all the frames and chunks in the computation. This provides strong generation quality but is unsuitable for streaming generation or long-context extension. Streaming autoregressive generation, however, models the conditional generation process as,

\begin{equation}
p_\theta(\mathbf{x}^{1:K} \mid \mathcal{C}^{1:K})
=
\prod_{k=1}^{K}
p_\theta(\mathbf{x}^{k} \mid \mathbf{x}^{<k}, \mathcal{C}^{\leq k}),
\end{equation}
where the video is divided into $K$ chunks, and $C$ represents the conditional contexts available . The causal architecture enables a KV caching mechanism to accelerate processing and progressive generation beyond a fixed offline clip length. Suppose we already have a strong bidirectional teacher, and converting the teacher to a causal student becomes the most challenging step. We formulate a generic framework to convert a bidirectional model to an autoregressive causal model. Specifically, EditStream introduces a recipe that keeps the teacher bidirectional while converting it into a causal student through a teacher-forced warm-up stage followed by unrolling distillation. The recipe can be generalized to any customized video training tasks.

\subsection{Teacher Adaptation with Block-wise Teacher Forcing Finetuning}

To better prepare the teacher model for subsequent block-wise distillation, we augment the standard full-sequence diffusion training with two additional training modes: \textbf{first-block cold-start training} and \textbf{prefix-suffix teacher forcing training}. 

\textbf{First-block cold-start training.} Let the latent video be divided into temporal blocks of \(B=3\) latent frames. With probability \(p_{\text{first}}\), we truncate all tensors to the first block and train the teacher to denoise only this short clip. This is because we assume the model must learn to synthesize the initial video segment well without any context history as the most challenging step.

\textbf{Prefix-suffix teacher forcing training.} We apply prefix-suffix teacher forcing with probability \(p_{\text{ps}}\). Instead of denoising the entire latent sequence, we uniformly sample one of the temporal blocks as the denoising target. The selected block is corrupted with noise at a sampled diffusion timestep, while all other blocks are replaced by their ground truth clean latents and assigned timestep zero. The training loss is computed only on the selected noisy block. In other words, the teacher is trained to predict the velocity of a single block conditioned on both clean prefix and clean suffix context. 

This training mixture yields about $p_{\text{first}}$ first-block short-clip training, $p_{\text{ps}}$ prefix-suffix teacher-forcing training, and ($1 - p_{\text{first}}-p_{\text{ps}}$ ) full-sequence training. The design has several advantages for distillation. First-block training directly strengthens the most challenging generation stage with a high training ratio. And the prefix-suffix conditioning provides a strong local denoising target for any block while preserving temporal consistency with \textit{surrounding} clean context. The mixture of the two training modes aligns the teacher’s training distribution with the student’s block-wise generation distribution well. Also, keeping a portion of standard full-sequence training preserves the teacher’s global video modeling ability, so the teacher remains a high-quality reference model while becoming better suited for downstream distillation.

Different from prior causal-teacher approaches \cite{zhu2026causalforcing, zhao2026causalforcingp, zheng2026causalrcm} that convert the teacher itself into a causal or streaming model, our teacher at this stage remains a full-context bidirectional diffusion model throughout training. We do not modify its attention mask. Instead, causality is only introduced at the level of the noising pattern and the loss mask. This is because causalizing a pretrained bidirectional video diffusion teacher can potentially reduce its modeling capacity and break the full-sequence distribution it was originally trained on. In contrast, our formulation preserves the teacher's native bidirectional generation ability and uses it as a high-quality target for block-wise supervision.

\subsection{Student Distillation Framework}

After obtaining a block-wise teacher-forced teacher, we distill it into a causal student for streaming video generation. The distillation framework contains two stages:
\begin{enumerate}
\item \textbf{Stage 1: Warm-up distillation}: a stable teacher-forced distillation stage as initialization, where the causal student is trained on single denoising blocks under clean prefix-suffix context from the teacher. 
\item \textbf{Stage 2: Unrolling distillation}: a full-sequence unrolling stage as post-training, where the causal student is autoregressively unrolled with KV caching and trained under its own generated history. 
\end{enumerate}

Training can transition smoothly from the first stage to the second after a small number of iterations.

\subsubsection{Stage 1: Warm-up Distillation}
The warm-up stage trains the causal student using a mixture of
first-block and prefix--suffix teacher-forcing updates with the VMM
objective defined above. The teacher \(T\) remains frozen and uses
bidirectional attention. The auxiliary model \(A\) is also bidirectional,
whereas the student \(S\) uses block-causal attention. Both the auxiliary
model and the student are initialized from the pretrained teacher
checkpoint.

We divide the latent video into temporal blocks, each containing \(B=3\) latent frames. Same as the teacher, at each iteration, one block is selected as the
active denoising block and all other blocks are kept at the ground-truth clean values
and their timesteps are set to zero. The active block is corrupted at a
sampled timestep \(t\), and the student performs a one-step ODE update
from \(t\) to a lower timestep \(s\). Let \(\mathbf{m}\) denote the
binary mask of the active block.

Since different blocks are evaluated at different timesteps, we define
the mixed-time video state as
\begin{equation}
\widetilde{\mathbf{x}}_{\ell}
=
(1-\mathbf{m})\odot\mathbf{x}_0
+
\mathbf{m}\odot\mathbf{x}_{\ell},
\qquad
\widetilde{\boldsymbol{\tau}}_{\ell}
=
\ell\,\mathbf{m},
\qquad
\ell\in\{t,s\}.
\end{equation}
Thus, only the active block is evaluated at timestep \(\ell\), while
all clean anchor blocks remain at timestep zero.

The auxiliary and student losses then become
\begin{equation}
\mathcal{L}_{A}
=
\E
\left[
\left\|
\mathbf{m}\odot
\left(
v_A\!\left(
\stopgrad(\widetilde{\mathbf{x}}_s),
\widetilde{\boldsymbol{\tau}}_s,
\mathcal{C}
\right)
-
\stopgrad(\bar v_S^{t\to s})
\right)
\right\|_2^2
\right],
\end{equation}
and
\begin{equation}
\mathcal{L}_{S}
=
\E
\left[
\left\|
\mathbf{m}\odot
\left(
\bar v_S^{t\to s}
-
\stopgrad\left(
\bar v_S^{t\to s}
+
v_T\!\left(
\widetilde{\mathbf{x}}_s,
\widetilde{\boldsymbol{\tau}}_s,
\mathcal{C}
\right)
-
v_A\!\left(
\widetilde{\mathbf{x}}_s,
\widetilde{\boldsymbol{\tau}}_s,
\mathcal{C}
\right)
\right)
\right)
\right\|_2^2
\right].
\end{equation}

In the prefix--suffix setting, the clean prefix and suffix blocks are
retained as anchors, while the loss mask \(\mathbf{m}\) selects only the
active noisy block. The bidirectional teacher and auxiliary model can
attend to both the clean prefix and suffix, whereas the block-causal
student can attend only to the clean prefix and the current active block.
Therefore, the clean suffix acts as privileged context for constructing
the bidirectional VMM correction, but it is not directly available to
the causal student. For first-block updates, all subsequent clean blocks are truncated, such
that the teacher, auxiliary model, and student process only the first
block. 

The warm-up stage provides a stable few-step autoregressive
initialization and empirically reduces block-boundary jitter. Since the
student is still conditioned on ground-truth clean anchors during this
stage, the warm-up objective does not by itself eliminate the
teacher-forcing exposure gap, which is further addressed by the
subsequent unrolling stage. More comparison with other initialization approaches can be found in the Appendix. 

\subsubsection{Stage 2: Unrolling Distillation}
\label{sec:unrolling_distillation}

\begin{figure*}[t]
    \centering
    \includegraphics[width=\textwidth]{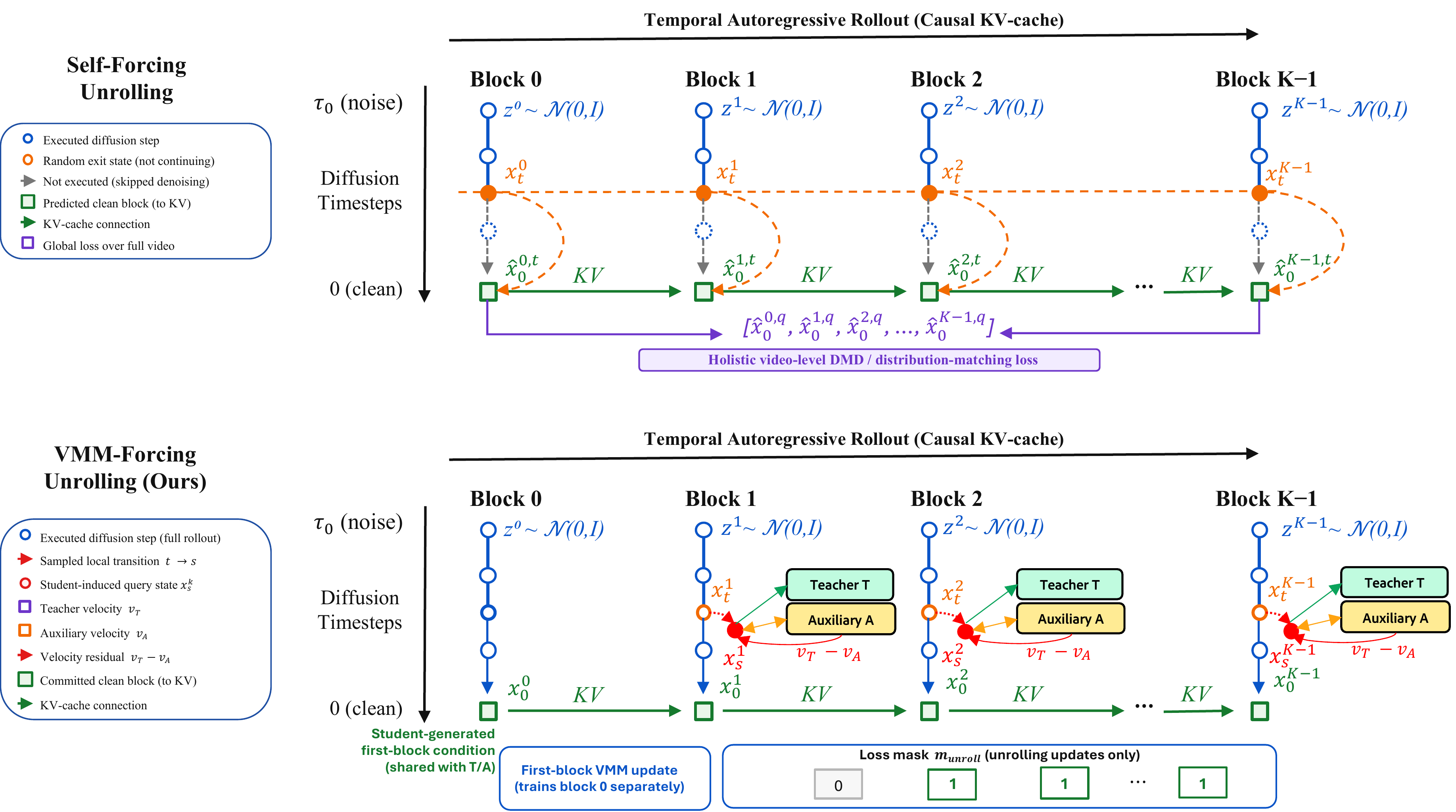}
    \caption{Overview of \textbf{VMM-forcing unrolling}. During training, we fully unroll streaming inference: the causal student generates each temporal block from noise with the target few-step schedule, commits the denoised block to the KV cache as history, and proceeds to the next block. At a sampled local transition for each selected block/rollout, we query the bidirectional teacher and auxiliary model at the student-induced states, and supervise the student by matching the teacher--auxiliary residual velocity moments under the same autoregressive context. The student-generated first-block condition is also fed into the teacher and auxiliary model.}
    \label{fig:unrolling}
\end{figure*}

The warm-up stage stabilizes local block generation, but the causal student still conditions on ground-truth clean context. As explained in self-forcing~\citep{huang2026self}, during
streaming inference, each block is conditioned on its previously generated and committed blocks. In this report, we introduce a new full-unrolling distillation approach to reduce the train-inference gap as post-training, named \textbf{VMM-Forcing unrolling}. The difference between VMM-Forcing unrolling and the previous self-forcing unrolling is illustrated in Figure \ref{fig:unrolling} and explained in the Appendix. 

During VMM-forcing unrolling, the student follows the same temporal autoregressive
execution and KV-cache mechanism used at inference. Let

\[
\tau_0 >  \cdots > \tau_{J-1} > \tau_{J} = 0
\]

denote the few-step inference schedule. For each block \(k\), the student
starts from Gaussian noise and follows its complete denoising trajectory:
\begin{equation}
\mathbf{x}_{\tau_0}^{k}
\sim
\mathcal{N}(\mathbf{0},\mathbf{I}),
\qquad
\mathbf{x}_{\tau_{j+1}}^{k}
=
\operatorname{Step}_{S}
\left(
\mathbf{x}_{\tau_j}^{k},
\tau_j,
\tau_{j+1};
\mathcal{H}_{S}^{<k}
\right),
\end{equation}
where \(\mathcal{H}_{S}^{<k}\) denotes the committed
student-generated history before block \(k\). After completing the full few-step trajectory, the resulting clean block
$\widehat{\mathbf{x}}_{0}^{k,S}
=\mathbf{x}_{\tau_J}^{k}$ is written into the causal KV cache at timestep zero:
\begin{equation}
\mathcal{H}_{S}^{\leq k}
=
\operatorname{Append}
\left(
\mathcal{H}_{S}^{<k},
\operatorname{KV}_{S}
\left(
\stopgrad(\widehat{\mathbf{x}}_{0}^{k,S}),
0
\right)
\right).
\end{equation}
The committed block then becomes the causal context for generating later
blocks.

During this full trajectory, while applying the VMM recipe, we sample one inference interval with source
timestep \(t\), and sample \(s<t\) inside that interval. Starting from the
corresponding student state \(\mathbf{x}_{t}^{k}\), we construct a local
VMM query state
\begin{equation}
\mathbf{x}_{s}^{k}
=
\mathbf{x}_{t}^{k}
+
(\sigma_s-\sigma_t)
v_S
\left(
\mathbf{x}_{t}^{k},
t,
\mathcal{H}_{S}^{<k}
\right).
\label{eq:unroll_vmm_state}
\end{equation}
VMM is evaluated at this intermediate state, while the block rollout
continues until timestep zero before its output is committed to the KV
cache. Thus, the state used for the local VMM correction and the state
used as context for subsequent blocks play different roles: the former
provides trajectory-local supervision, whereas the latter reproduces the
student's inference-time committed output.

The unrolling stage alternates between two types of updates and one special condition:
\begin{itemize}
    \item \textbf{Update 1: First-block VMM update.}
    We apply the non-unrolled first-block VMM path introduced in the
    warm-up stage. This continuously preserves the cold-start generation
    quality of block \(0\).

    \item \textbf{Update 2: Full-unrolling update.}
    We roll out all temporal blocks using the complete few-step student
    sampler and causal KV cache. Block \(0\) is treated as committed
    context and masked out from the unrolling loss, while later blocks
    are optimized under student-generated history.
    \item \textbf{Shared student-generated first-block context:} For an unrolling update, the committed first block generated by the student and in the KV cache is  also fed into the teacher and auxiliary model as their first-block condition. This ensures that the student, teacher, and auxiliary model construct their predictions under the same initial scene and appearance. Because block \(0\) has no autoregressive history and remains explicitly optimized by the separate first-block VMM path, it provides a relatively reliable shared anchor for downstream unrolling. This does not assume that the first block is error-free, but it avoids the history-induced exposure bias that affects later blocks.
    
\end{itemize}

Let
\(\bar v_{S}^{k,t\rightarrow s}\) denote the student transition velocity
associated with the sampled local exit step for block \(k\). The teacher
and auxiliary velocities are evaluated at the same assembled
student-induced state:
\begin{equation}
v_T^k
=
v_T
\left(
\mathbf{x}_s^k,
s,
\mathcal{C}^{k}
\right),
\qquad
v_A^k
=
v_A
\left(
\mathbf{x}_s^k,
s,
\mathcal{C}^{k}
\right),
\end{equation}
where \(\mathcal{C}^{k}\) includes
the committed student first block. The auxiliary model is trained with
\begin{equation}
\mathcal{L}_{A}^{\mathrm{unroll}}
=
\mathbb{E}_{k,t,s}
\left[
\left\|
\mathbf{m}_{\mathrm{unroll}}
\odot
\left(
v_A^k
-
\stopgrad
\left(
\bar v_{S}^{k,t\rightarrow s}
\right)
\right)
\right\|_2^2
\right],
\end{equation}
and the student is trained with
\begin{equation}
\mathcal{L}_{S}^{\mathrm{unroll}}
=
\mathbb{E}_{k,t,s}
\left[
\left\|
\mathbf{m}_{\mathrm{unroll}}
\odot
\left(
\bar v_{S}^{k,t\rightarrow s}
-
\stopgrad
\left(
\bar v_{S}^{k,t\rightarrow s}
+
v_T^k
-
v_A^k
\right)
\right)
\right\|_2^2
\right],
\label{eq:vmm_unrolling_loss}
\end{equation}
where \(\mathbf{m}_{\mathrm{unroll}}\) selects the downstream blocks
optimized during unrolling and masks out the committed first block.

Overall, the hybrid stage combines stable first-block learning with
closed-loop trajectory refinement. The separate first-block path
maintains cold-start quality, while full unrolling exposes later blocks
to the causal context distribution produced by the deployed student
sampler. The teacher and auxiliary model then provide VMM corrections at
intermediate states encountered along these inference-like trajectories.

\section{Data Construction}

A unified streaming video editing model must support both general video generation and a wide range of editing behaviors, including text-to-video generation, image-to-video, instruction-guided video-to-video editing, editing propagation, reference-guided video-to-video editing, object addition/removal, egocentric manipulation, visual-effect removal, and camera pose control. We therefore train EditStream on a heterogeneous mixture of generative data, paired editing data, and physically grounded video data. All videos are sampled as 81-frame clips and resized into 720p aspect-ratio buckets, mainly \(704\times1280\) and \(1280\times704\). Table~\ref{tab:data_mix} summarizes the training mixture. The ratios denote sampling ratios during training.
\begin{table*}[htbp]
\centering
\caption{\textbf{Training Data Mixture.}
Sources, dataset scales, and sampling ratios used to train EditStream.}
\label{tab:data_mix}
\resizebox{\textwidth}{!}{
\begin{tabular}{lll|c}
\toprule
Category
& Source Family
& Scale
& Sampling Ratio \\
\midrule

\tablegroupheader{4}{Video Generation Data}

Text-to-Video
& LTX2.3-video, OpenVid-1M \cite{nan2025openvid}
& 205K + 1.02M ($\sim$1.23M clips)
& 21.875\% \\

Text/Image-to-Video
& LTX2.3-video, OpenVid-1M \cite{nan2025openvid}
& 205K + 1.02M ($\sim$1.23M clips)
& 21.875\% \\

\midrule
\tablegroupheader{4}{Video Editing and Control Data}

Image Editing / Propagation
& Pico-style image editing \cite{qian2026pico}
& $\sim$328K image pairs
& 9.375\% \\

General Video Editing / Propagation
& Curated paired video editing
& $\sim$37K pairs
& 6.25\% \\

Egocentric Video Editing / Propagation
& EgoEdit \cite{li2026egoedit}
& $\sim$94K pairs
& 6.25\% \\

Video Effect Removal / Propagation
& EffectErase \cite{fu2026EffectErase}
& $\sim$96K pairs
& 6.25\% \\

Reference-Guided Video Editing
& Kiwi RefVIE \cite{lin2026kiwi}
& $\sim$476K pairs
& 6.25\% \\

Human--Object Interaction / Propagation
& HUMOTO \cite{lu2025humoto}
& $\sim$21K clips
& 9.375\% \\

Synthetic Object Editing / Propagation
& Kubric \cite{greff2022kubric}
& $\sim$50K clips
& 6.25\% \\

Explicit Camera Pose Control
& MultiCamVideo \cite{bai2025recammaster}
& $\sim$57K clips
& 6.25\% \\

\bottomrule
\end{tabular}
}
\end{table*}

\textbf{Generation and Image-to-Video Data.} We allocate roughly half of the training mixture to general video generation and image-conditioned video generation. Specifically, text-to-video samples are from LTX2.3-video and OpenVid-1M \cite{nan2025openvid}, and text/image-to-video samples are from the same sources. For the LTX2.3-video dataset, we run the inference of LTX2.3~\citep{hacohen2026ltx}  on user prompts introduced in Self-forcing. This regularization preserves the base model's general video prior and prevents the unified editor from overfitting to paired editing data only.

\textbf{General Editing and Propagation Data.} We use paired video editing data to teach instruction-following transformations. This includes self-collected general source-target video pairs and Pico \cite{qian2026pico} image editing pairs converted into short video training examples with smooth translation augmentation. For propagation training, the model receives either an edited first frame or a weak propagation prompt and learns to propagate the edit to the entire video results. These propagation samples are important for interactive workflows, when users want to edit a single frame, and manipulate the video editing based on the first frame. 

\textbf{Reference-Guided Editing Data.} We include KiwiEdit \cite{lin2026kiwi} reference-guided editing data, containing roughly 476K source-target-reference triplets. These samples provide a source video, an editing instruction, and an additional reference image. They train the model to transfer subject or background from the reference image to the target video while preserving the consistency of the source video.

\textbf{Physically Grounded Editing Data.} Inspired by VOID \cite{motamed2026void}, to improve physical consistency, we include several rendering data sources. HUMOTO \cite{lu2025humoto} contains approximately 21K human-object interaction clips paired with corresponding object-present and object-absent videos. We use it for object addition, object removal, and edit propagation.

To improve the diversity of this dataset, we curate indoor and outdoor 3D scenes from Fab \cite{fab}, Sketchfab \cite{sketchfab}, BlenderKit \cite{blenderkit}, and other asset sources. For each selected HUMOTO motion, we manually identify valid sub-sequences and target interaction objects. Each rendered sample is specified by a scene, a HUMOTO sub-sequence, a target object, and a camera trajectory. We render two temporally synchronized videos with identical scene layout, camera motion, human motion, lighting, and appearance: one containing the target object and one with the target object removed. To increase viewpoint diversity, we render third-person, first-person, and object-centric views.

We apply Gemini-based \cite{comanici2025gemini} VLM filtering to reject low-quality pairs showing penetration, floating artifacts, unclear human-object interactions, heavy occlusion, or strange viewpoints. To improve visual diversity, we randomize human and object appearances across samples while keeping the two videos in each pair appearance-consistent.

We generate paired add/remove editing instructions using Qwen3-VL-32B \cite{bai2025qwen3} conditioned on the rendered videos and the original HUMOTO captions. For object addition, we provide two instruction variants: a concise object-centric instruction without motion details, and an interaction-aware instruction that explicitly describes the human-object interaction. For video-to-video editing training, we utilize the interaction-aware instruction, while for propagation, we only describe the object. 

Kubric-style \cite{greff2022kubric} synthetic object editing contributes around 50K clips and is used for both object removal and object addition. We also use EgoEdit \cite{li2026egoedit} for egocentric object replacement and EffectErase \cite{fu2026EffectErase} for removing visual effects and their induced artifacts. 

\textbf{Camera Pose Control Data.} We construct camera-control training data from the MultiCamVideo \cite{bai2025recammaster} dataset, which provides synchronized multi-view videos with 10 cameras per dynamic scene. Each camera sequence contains 81 frames, and the dataset covers four focal-length/aperture groups: f18-aperture10, f24-aperture5, f35-aperture2.4, and f50-aperture2.4. For each focal group, we process the first 512 scenes by default. Within each scene, we deterministically sample 30 ordered source-target camera pairs from all possible camera pairs.

For every selected scene, we run VGGT-Omega \cite{wang2026vggt} once using all 10 camera videos and all 81 frames. VGGT-Omega predicts per-frame camera poses, intrinsics, and depth maps for all views in a shared reconstruction space. For each sampled source-target pair, we unproject the predicted source-view RGB-D sequence into a 4D point cloud and render it into the VGGT-Omega-estimated target camera trajectory. This produces a point-cloud-rendered video that is geometrically aligned with the target view as closely as possible.

Each training sample consists of the source video, the point-cloud render, the target video, the point-cloud alpha mask, and the corresponding camera metadata. The target video serves as the supervision signal, while the point-cloud render and alpha mask provide geometric conditioning for the video model. Since monocular or learned 3D reconstruction can be imperfect, VGGT-Omega may produce inaccurate poses or depths in some scenes. We therefore compute alignment-related quality statistics, including valid-render coverage and pixel-alignment metrics, and use them to filter generated samples before forming the final training set. Captions are generated with Qwen3-VL-32B-Instruct \cite{bai2025qwen3} from uniformly sampled target-video frames, producing camera-agnostic scene descriptions used as text conditioning.

Overall, the final dataset balances general generation, image-conditioned generation, instruction editing, propagation, reference guidance, physical object editing, egocentric editing, effect removal, and camera control. This balance is essential for a unified streaming editor: the same model must preserve strong generative quality while handling diverse, temporally coherent editing tasks.

\section{Experiments}

\subsection{Quantitative and Qualitative Results}

\begin{table}[htbp]
\centering
\caption{\textbf{Text-to-Video} Results: VBench text-to-video benchmark. Latency and throughput are both end-to-end performance. Time-to-first-output denotes the wall-clock time until the first decoded RGB block becomes available at batch size 1. Measurements are conducted in one single H100 GPU. }
\label{tab:vbench_t2v}
\resizebox{\textwidth}{!}{%
\renewcommand{\arraystretch}{1.15}
\begin{tabular}{lcccc|cccc|ccc}
\toprule
\begin{tabular}[c]{@{}c@{}}Model\end{tabular}
& \begin{tabular}[c]{@{}c@{}}\#Params\end{tabular}
& \begin{tabular}[c]{@{}c@{}}Resolution\end{tabular}
& \begin{tabular}[c]{@{}c@{}}Latency:\\ time-to-first-output\\ (s)\end{tabular}
& \begin{tabular}[c]{@{}c@{}}Steady-state\\ Throughput\\ (FPS)\end{tabular}
& \begin{tabular}[c]{@{}c@{}}Dynamic\\ Degree $\uparrow$\end{tabular}
& \begin{tabular}[c]{@{}c@{}}Color $\uparrow$\end{tabular}
& \begin{tabular}[c]{@{}c@{}}Human\\ Action $\uparrow$\end{tabular}
& \begin{tabular}[c]{@{}c@{}}Multiple\\ Objects $\uparrow$\end{tabular}
& \begin{tabular}[c]{@{}c@{}}Quality\\ Score $\uparrow$\end{tabular}
& \begin{tabular}[c]{@{}c@{}}Semantic\\ Score $\uparrow$\end{tabular}
& \begin{tabular}[c]{@{}c@{}}Total\\ Score $\uparrow$\end{tabular} \\
\midrule

\vbenchgroupheader{Offline Models}
Wan2.2 5B TI2V Base \cite{wan2025}
& 5B 
& 1280$\times$704
& 72.30
& 1.12
& \underline{63.61}
& \textbf{89.52}
& \underline{97.20}
& \textbf{84.88}
& \underline{84.04}
& \textbf{81.86}
& \underline{83.60} \\

\textbf{EditStream Teacher FT} (ema)
& 5B 
& 1280$\times$704
& 72.30
& 1.12
& \textbf{67.78}
& \underline{86.15}
& \textbf{97.80}
& \underline{84.15}
& \textbf{84.73}
& \underline{81.31}
& \textbf{84.05} \\

\midrule
\vbenchgroupheader{Streaming Models}

\textbf{EditStream-AR} (ODE Init. + Self-Forcing Unrolling \cite{huang2026self})
& 5B
& 1280$\times$704
& 0.90 
& 10.56
& 48.06
& 82.72
& 96.80
& \textbf{89.54}
& 84.09
& \textbf{81.95}
& 83.67 \\

\textbf{EditStream-AR (VMM Init. + Self-Forcing Unrolling)}
& 5B
& 1280$\times$704
& 0.90 
& 10.56
& \underline{66.67}
& \underline{84.45}
& \underline{97.40}
& \underline{84.91}
& \underline{85.03}
& 80.52
& \underline{84.13} \\

\textbf{EditStream-AR (VMM Init. + VMM-Forcing Unrolling)}
& 5B
& 1280$\times$704
& 0.90 
& 10.56
& \textbf{79.17}
& \textbf{87.82}
& \textbf{98.00}
& 83.26
& \textbf{85.36}
& \underline{80.64}
& \textbf{84.42} \\

\bottomrule
\end{tabular}
}
\end{table}

\begin{figure*}[htbp]
    \centering
    \includegraphics[width=\textwidth]{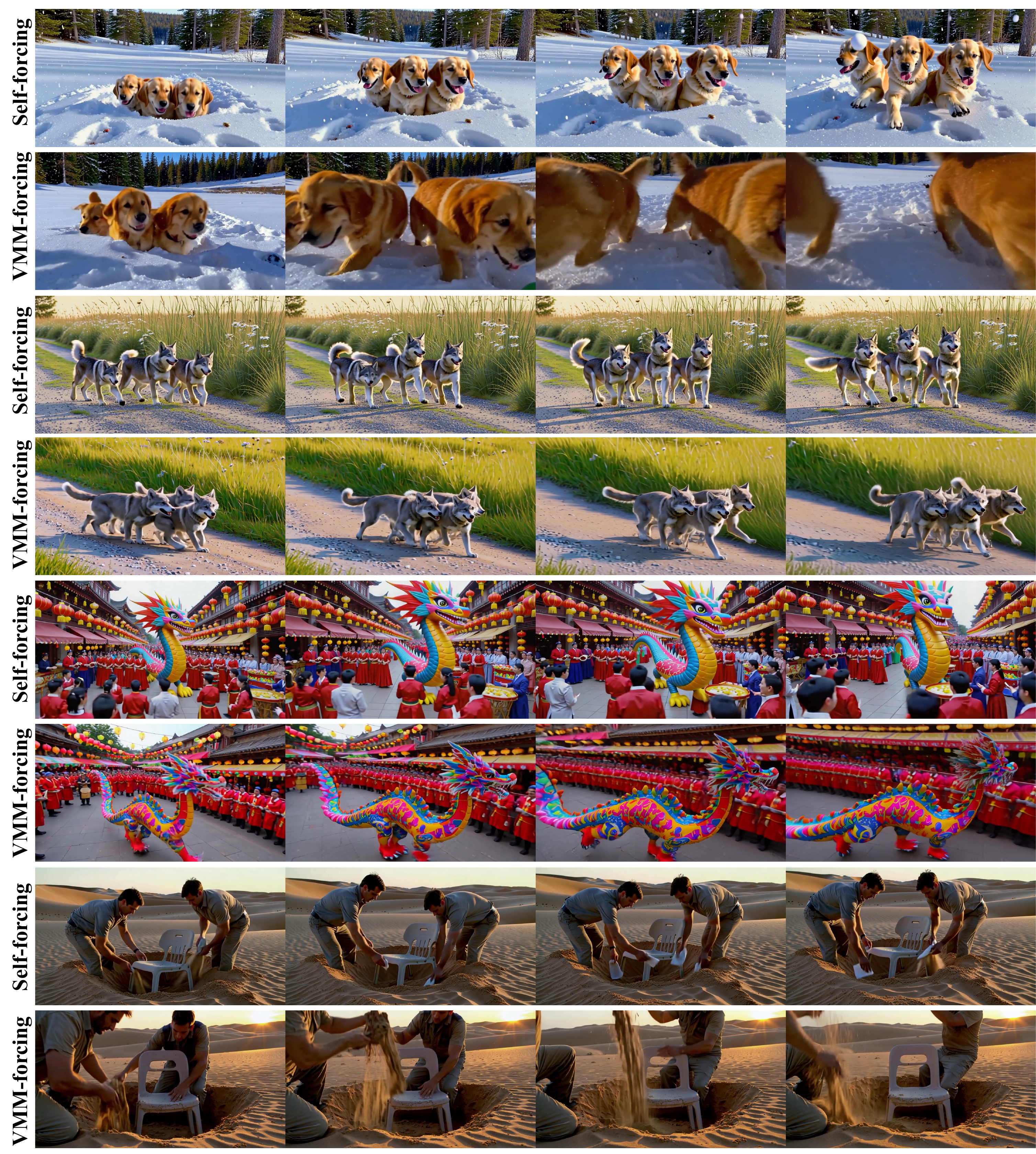}
    \caption{Qualitative Text-to-Video (T2V) results on frequent user prompts. We compare the Self-forcing conversion and the proposed VMM-forcing pipeline. VMM-forcing produces richer motion dynamics and scene evolution while maintaining texture realism and temporal coherence, whereas Self-forcing often yields over-sharpened, plastic-like appearances and reduced temporal diversity.}
    \label{fig:t2v}
\end{figure*}

\textbf{Text-to-Video (T2V) Results.} 
Table~\ref{tab:vbench_t2v} reports the VBench evaluation on text-to-video generation. Fine-tuning the teacher slightly improves the overall score over the Wan2.2 base model (84.05 vs. 83.60). VMM-forcing Unrolling achieves the best overall score (84.42), outperforming both the teacher and the self-forcing baseline. The improvement is primarily driven by stronger dynamic degree (79.17), color quality (87.82), human action (98.00), and perceptual quality (85.36). The results suggest that VMM-forcing retains greater motion and scene diversity than the evaluated Self-Forcing baselines. At 720p, EditStream produces \textbf{10.56} FPS under our evaluation setup on one single H100 GPU. If we parallelize VAE decoding with DiT denoising on two GPUs, the throughput can be increased to \textbf{16.10} FPS. It enables low-latency progressive output, although it does not yet reach real-time playback speed at this resolution and requires more industry-level optimization and better inference hardware. Pixel-aligned editing tasks use the same causal backbone and have similar measured runtime. Token-conditioned tasks introduce only a small task-dependent overhead.

To further investigate why the \texttt{multiple\_objects} score of the Self-forcing baseline is higher than that of VMM-forcing, we hypothesize that this is because the generated videos have stronger motion, causing multiple objects to move out of the frame. To verify this hypothesis, we re-scored the 410 results using the continuous motion score from the RAFT-based optical flow in VBench. The results show that the \texttt{multiple\_objects} score of VMM-forcing is negatively correlated with motion: among all results with a score of 0, the average motion is 16.8, while among all videos with a score of 1, the average motion is only 8.5. We then compared the results with Self-forcing on a sample-by-sample basis. Among the 92 prompts where VMM-forcing loses to Self-forcing, 69.6\%
of the cases have stronger motion, while among the 38 prompts where VMM-forcing wins, this proportion is only 57.9\%. This shows that samples with stronger motion are concentrated in the cases where VMM-forcing loses. We manually inspected several examples, and for some prompts, such as ``an airplane and a train,'' the camera moves or zooms, causing objects to leave the frame. Of course, there are also some cases where the detector fails: although we visually observe that the objects are generated completely and remain visible, they still receive a failing score, showing that the detector may have some limitations.

Figure~\ref{fig:t2v} provides qualitative comparisons between the proposed VMM-forcing pipeline and the self-forcing baseline using prompts sampled from the MovieGen \cite{polyak2024movie} prompt set. Across a variety of common user prompts, VMM-forcing consistently generates videos with richer scene evolution and more dynamic camera and object motions. For example, in the multi-animal examples, the movements and interactions of the dogs and wolves are more diverse and realistic than those produced by self-forcing. In contrast, the ODE-based initialization adopted by the self-forcing baseline tends to produce relatively static camera trajectories and limited scene evolution. We also observe signs of mode collapse in the generated appearance, where outputs become overly sharp but exhibit artificial, plastic-like textures. These qualitative observations are consistent with the quantitative improvements in dynamic degree and perceptual quality reported in Table~\ref{tab:vbench_t2v}, suggesting that VMM-forcing better preserves the stochasticity and temporal diversity of the original diffusion process while maintaining high visual fidelity.

\begin{table}[htbp]
\centering
\caption{\textbf{Video-to-Video Editing}: Comparison on EditVerse benchmark for video-to-video editing.
Higher values indicate better performance for all metrics. We follow EditVerse to use GPT-4o as the VLM judge model. }
\label{tab:editverse}
\resizebox{\textwidth}{!}{
\begin{tabular}{lcccccc}
\toprule
Model
& CLIP Temp. $\uparrow$
& DINO Temp. $\uparrow$
& Frame-Text Align. $\uparrow$
& Video-Text Align. $\uparrow$
& Pick Score $\uparrow$
& Video Quality $\uparrow$ \\
\midrule

\groupheader{Commercial Models}

Runway Aleph
& \underline{0.9888}
& \underline{0.9849}
& \textbf{28.30}
& \underline{25.33}
& \underline{20.36}
& \underline{7.64} \\

Kling-O1
& \textbf{0.9926}
& \textbf{0.9911}
& \underline{28.24}
& \textbf{25.72}
& \textbf{20.59}
& \textbf{7.93} \\

\midrule
\groupheader{Offline Models}

InsV2V \cite{cheng2024consistent}
& 0.9720
& 0.9679
& 25.84
& 23.07
& 19.55
& 5.07 \\

LucyEditDev \cite{decart2025lucyedit}
& 0.9852
& 0.9839
& 26.30
& 23.51
& 19.54
& 5.87 \\

STDF \cite{yatim2024space}
& 0.9635
& 0.9612
& 26.18
& 23.59
& 19.74
& 4.55 \\

Senorita-2M \cite{zi2026senorita}
& 0.9804
& 0.9810
& 27.24
& 24.64
& 19.75
& 6.83 \\

TokenFlow \cite{geyer2024tokenflow}
& 0.9867
& 0.9880
& 26.63
& 24.26
& 20.01
& 5.34 \\

VACE \cite{jiang2025vace}
& \underline{0.9896}
& \underline{0.9884}
& 27.07
& 24.40
& \underline{20.06}
& 6.33 \\

EditVerse \cite{ju2025editverse}
& 0.9859
& 0.9859
& \textbf{27.89}
& \textbf{25.66}
& \textbf{20.12}
& \textbf{7.67} \\

\textbf{EditStream Teacher}
& \textbf{0.9917}
& \textbf{0.9901}
& \underline{27.49}
& \underline{25.31}
& \textbf{20.12}
& \underline{7.52} \\

\midrule
\groupheader{Streaming Models}

SANA Streaming \cite{zhao2026sana}
& 0.9889
& 0.9869
& 26.91
& 24.55
& 19.80
& 5.95 \\

LiveEdit \cite{wang2026liveedit}
& 0.9899
& \underline{0.9896}
& 25.91
& 23.28
& 19.45
& 6.52 \\

\textbf{EditStream-AR} (ODE Init. + Self-Forcing Unrolling \cite{huang2026self})
& 0.9903
& 0.9892
& 26.81
& 24.68
& 19.98
& 7.01 \\

\textbf{EditStream-AR (VMM Init. + Self-Forcing Unrolling)}
& \textbf{0.9914}
& \textbf{0.9897}
& \underline{27.03}
& \textbf{25.22}
& \textbf{20.16}
& \underline{7.25} \\

\textbf{EditStream-AR (VMM Init. + VMM-Forcing Unrolling)}
& \underline{0.9911}
& 0.9893
& \textbf{27.05}
& \underline{24.95}
& \underline{20.01}
& \textbf{7.35} \\

\bottomrule
\end{tabular}
}
\end{table}
\begin{figure*}[htbp]
    \centering
    \includegraphics[width=\textwidth]{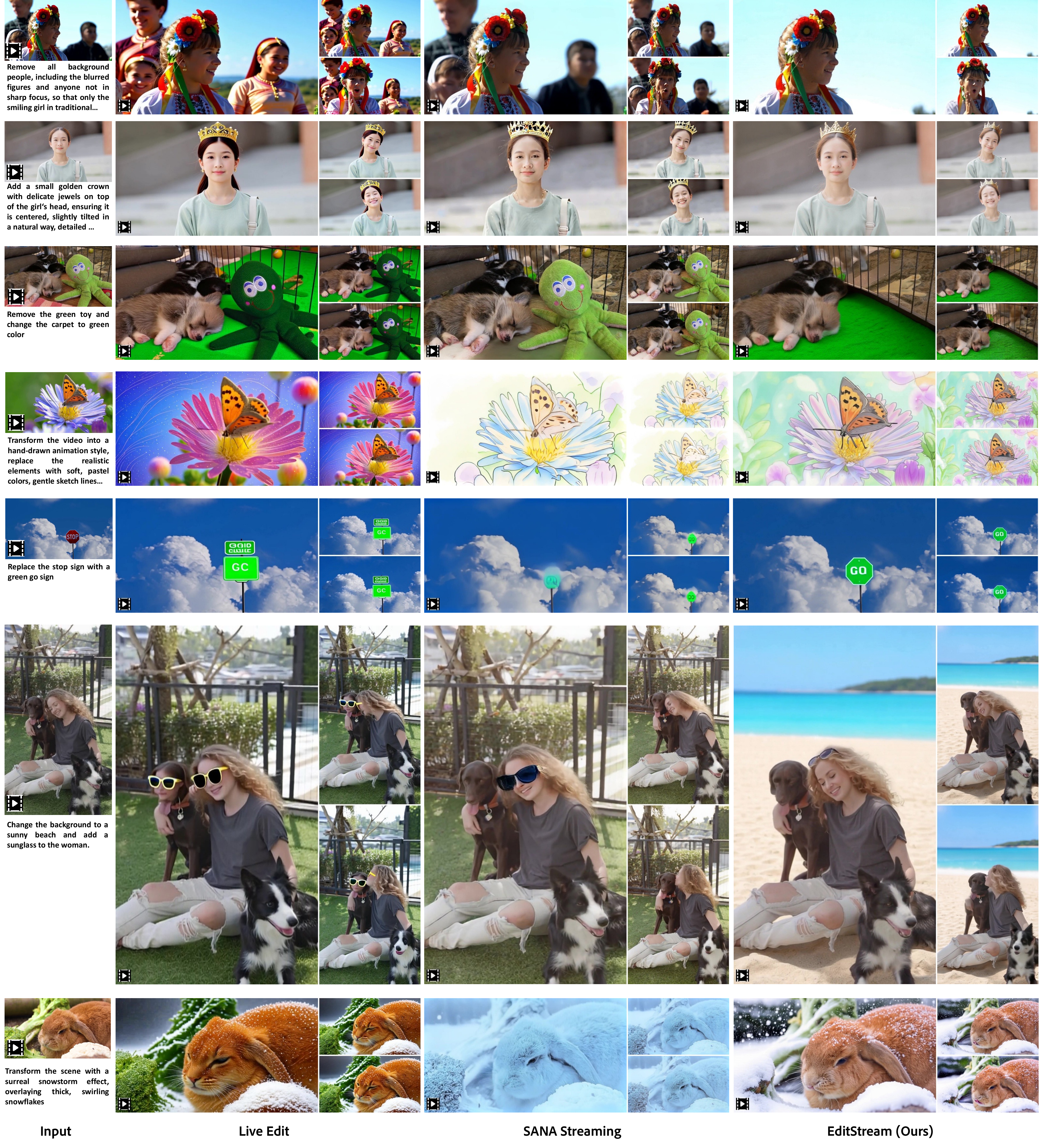}
    \caption{Video-to-Video (V2V) editing results on EditVerse Benchmark compared with other streaming video editing baselines.  Across a wide range of editing tasks, including object addition, removal, replacement, style transfer, text rendering, and complex compositional edits requiring precise pixel alignment and instruction following, EditStream consistently exhibits stronger semantic understanding, higher generation fidelity, and better color quality.}
    \label{fig:v2vresults}
\end{figure*}
\textbf{Video-to-Video (V2V) Editing Results.} 
Table~\ref{tab:editverse} reports video-to-video editing results on subset of EditVerse Benchmark. We select the categories of pixel-aligned video editing tasks forming a set of 110 testing samples. The table shows that the EditStream teacher achieves strong temporal consistency while remaining competitive on text alignment and perceptual quality. The streaming student variants preserve most of the teacher's performance after causal conversion. VMM-forcing based conversion pipeline achieves the highest video quality score judged by VLM. Compared with baseline streaming video editing works and baseline distillation scheme, EditStream achieves the highest VLM-based video-quality score among the evaluated streaming baselines.

In Figure \ref{fig:v2vresults}, qualitative comparisons against LiveEdit \cite{wang2026liveedit} and SANA Streaming \cite{zhao2026sana} further demonstrate the strengths of EditStream. Across a wide range of editing tasks, including object addition, removal, replacement, style transfer, text rendering, and complex compositional edits requiring precise pixel alignment and instruction following, EditStream consistently exhibits stronger semantic understanding, higher generation fidelity, and better color quality. However, we emphasize that these comparisons should be interpreted with caution, as differences in training data composition, model capacity, and the inclusion of image editing data during training can all affect editing quality. Consequently, it is difficult to attribute the observed performance gap solely to the proposed streaming architecture or distillation strategy. Therefore, these results demonstrate the effectiveness of the overall system design, including the training data, teacher training procedure, and the distillation strategy for preserving the teacher's editing capabilities. The contribution of each individual component cannot be cleanly disentangled from qualitative comparisons alone. 
\begin{table}[htbp]
\centering
\caption{\textbf{Reference-Guided Video-to-Video Editing}:
Comparison on RefVIE-Bench with Gemini-3-Flash as the judge model.}
\label{tab:refvie}
\resizebox{\textwidth}{!}{
\begin{tabular}{lccccccccc}
\toprule
\multirow{2}{*}{Model}
& \multicolumn{4}{c}{Subject Reference}
& \multicolumn{4}{c}{Background Reference}
& \multirow{2}{*}{Overall $\uparrow$} \\
\cmidrule(lr){2-5}
\cmidrule(lr){6-9}
&
\shortstack{Identity\\Consist. $\uparrow$}
&
\shortstack{Temporal\\Consist. $\uparrow$}
&
\shortstack{Physical\\Consist. $\uparrow$}
&
\shortstack{Subj.\\Avg. $\uparrow$}
&
\shortstack{Reference\\Similarity $\uparrow$}
&
\shortstack{Matting\\Quality $\uparrow$}
&
\shortstack{Video\\Quality $\uparrow$}
&
\shortstack{BG\\Avg. $\uparrow$}
& \\
\midrule

\refviegroupheader{Commercial Models}

Runway Aleph
& \underline{3.79}
& \underline{3.65}
& \underline{3.58}
& \underline{3.67}
& \underline{3.33}
& \underline{2.81}
& \underline{2.58}
& \underline{2.91}
& \underline{3.29} \\

Kling-O1
& \textbf{4.75}
& \textbf{4.66}
& \textbf{4.60}
& \textbf{4.67}
& \textbf{3.95}
& \textbf{3.21}
& \textbf{2.75}
& \textbf{3.30}
& \textbf{3.99} \\

\midrule

\refviegroupheader{Offline Models}

UniVideo \cite{wei2025univideo}
& \underline{4.17}
& \underline{3.79}
& \underline{3.59}
& \underline{3.85}
& 3.13
& 2.55
& 2.28
& \underline{2.65}
& \underline{3.44} \\

ReCo-Ref \cite{zhang2025region}
& 3.65
& 2.95
& 2.70
& 3.10
& \underline{3.35}
& \textbf{2.65}
& \underline{2.38}
& \textbf{2.79}
& 3.00 \\

Kiwi-Edit (Instruct-Reference)
& 3.51
& 2.96
& 2.91
& 3.13
& \textbf{3.40}
& \underline{2.58}
& \textbf{2.40}
& \textbf{2.79}
& 2.96 \\

Kiwi-Edit (Reference Only)
& \textbf{4.28}
& 3.61
& 3.55
& 3.81
& 3.33
& 2.35
& 2.08
& 2.58
& 3.40 \\

\textbf{EditStream Teacher}
& 4.13
& \textbf{3.96}
& \textbf{3.94}
& \textbf{4.01}
& 3.20
& 2.43
& 2.10
& 2.58
& \textbf{3.53} \\

\midrule

\refviegroupheader{Streaming Models}


\textbf{EditStream-AR} (ODE Init. + Self-Forcing Unrolling \cite{huang2026self})
& 3.49
& 3.30
& \underline{3.21}
& 3.33
& \underline{3.38}
& \textbf{2.78}
& \underline{2.43}
& \underline{2.86}
& 3.18 \\

\textbf{EditStream-AR (VMM Init. + Self-Forcing Unrolling)}
& \textbf{4.50}
& \textbf{4.10}
& \textbf{3.90}
& \textbf{4.17}
& 3.35
& \underline{2.68}
& \textbf{2.48}
& 2.83
& \textbf{3.72} \\

\textbf{EditStream-AR (VMM Init. + VMM-Forcing Unrolling)}
& \underline{4.41}
& \underline{4.03}
& \textbf{3.90}
& \underline{4.11}
& \textbf{3.45}
& \textbf{2.78}
& \textbf{2.48}
& \textbf{2.90}
& \underline{3.71} \\

\bottomrule
\end{tabular}
}
\end{table}

\begin{figure*}[htbp]
    \centering
    \includegraphics[width=\textwidth]{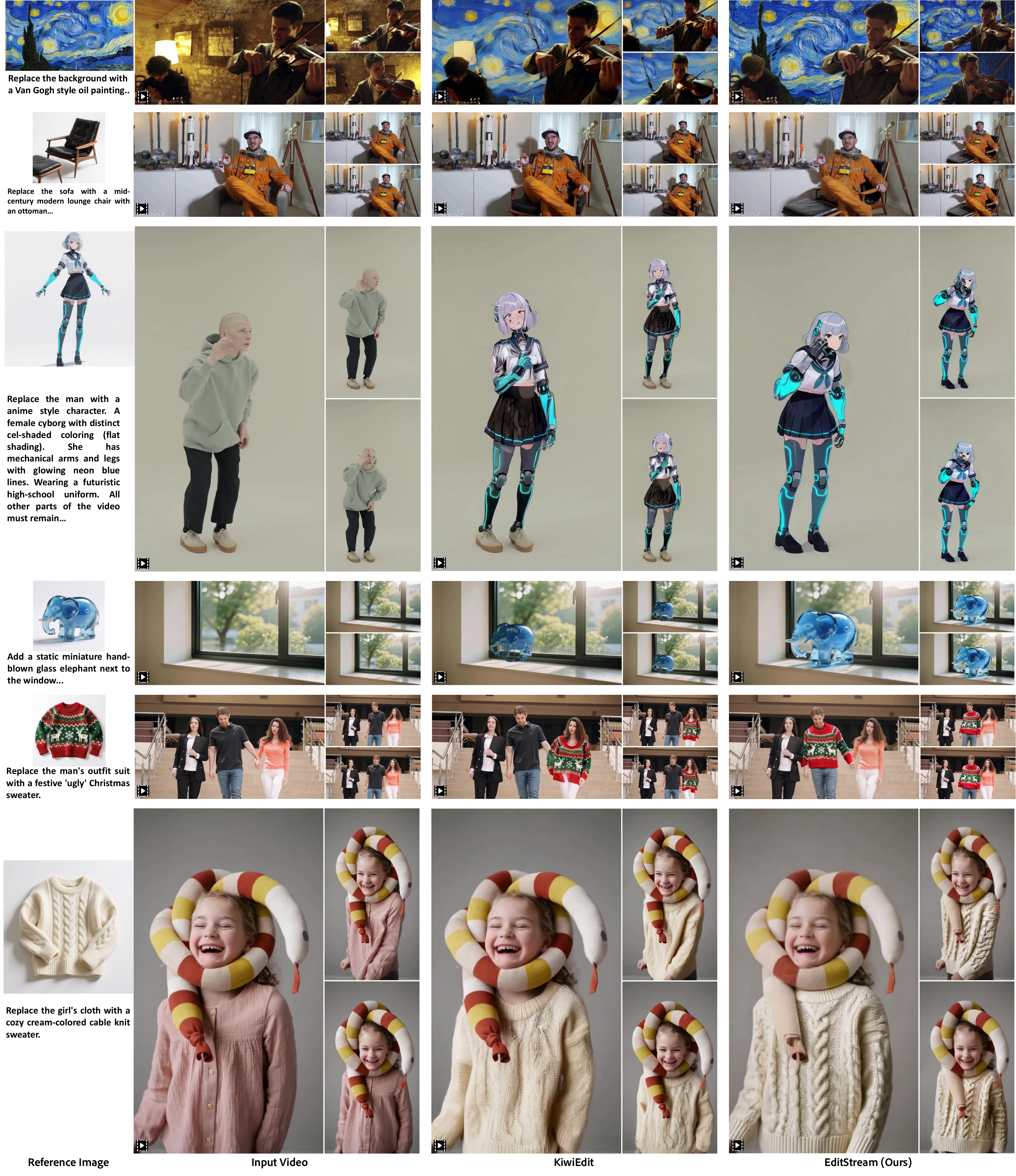}
    \caption{Reference Video-to-Video (refV2V) results on RefVIE Benchmark. Qualitatively, EditStream (online streaming model) consistently outperforms Kiwi-Edit (offline model) in preserving the reference subject while generating more realistic edited videos. Compared with the baseline, it produces more faithful identity preservation, more appropriate object scale, more coherent scene composition with the reference background, and richer pixel-level details.}
    \label{fig:refeditresults}
\end{figure*}

\textbf{Reference-Guided Video-to-Video (Ref-V2V) Editing Results.} 
Figure~\ref{fig:refeditresults} presents qualitative comparisons between EditStream and Kiwi-Edit \cite{lin2026kiwi}, where both methods are trained on the same reference-guided video dataset from KiwiEdit. Table~\ref{tab:refvie} shows that the EditStream teacher achieves the highest overall score (3.53) among open-source methods. It achieves the best subject-reference consistency (4.01) which demonstrates strong identity preservation and temporal coherence. After causal conversion and distillation, the streaming version maintains competitive performance. The newly proposed conversion pipeline also achieves better performance than the original ODE-initialized Self-Forcing pipeline.

Qualitatively, EditStream consistently outperforms Kiwi-Edit in preserving the reference subject while generating more realistic edited videos. Compared with the baseline, it produces more faithful identity preservation, more appropriate object scale, more coherent scene composition with the reference background, and richer pixel-level details. We attribute these improvements to the unified framework, which shares representations across multiple video editing tasks and enables effective transfer of editing capabilities to the reference-guided setting.  Finally, EditStream performs reference-guided editing in a low-latency streaming manner, substantially reducing inference latency while maintaining competitive editing quality.

\begin{table*}[htbp]
\centering
\caption{\textbf{Camera Pose Control}: Comparison on the V2VCameraPose benchmark (472 samples). The first four columns are VBench dimensions computed purely from the video itself (no prompt/text dependence); the last four evaluate the quality of content hallucinated into the disoccluded (never-observed) regions produced by reprojecting each sample's monocular point cloud onto the target camera trajectory, and how faithfully the observed (point-cloud-hit) region is preserved. Higher values indicate better performance for all metrics.}
\label{tab:camerapose}
\resizebox{\textwidth}{!}{
\begin{tabular}{lcccccccc}
\toprule
Model
& Subj. Cons. $\uparrow$
& Back. Cons. $\uparrow$
& Temp. Flicker $\uparrow$
& Motion Smooth. $\uparrow$
& Valid SSIM $\uparrow$
& Hole Complete. $\uparrow$
& Hole Temp. Stab. $\uparrow$
& Gated Hole Fill Qual. $\uparrow$ \\
\midrule

\rowcolor{gray!15}
\multicolumn{9}{l}{\textbf{Offline Models}} \\
ReCamMaster \cite{bai2025recammaster}
&\textbf{89.23}
&91.18
&\textbf{96.44}
&\textbf{98.65}
&0.388
&\textbf{96.96}
&82.07
&22.96 \\
ViewCrafter \cite{yu2024viewcrafter}
&86.86
&91.31
&93.26
&96.12
&0.407
&96.37
&\underline{86.43}
&27.73 \\
GCD \cite{van2024generative}
&84.03
&90.42
&92.39
&95.87
&0.315
&96.60
&83.71
&21.72 \\
TrajCrafter  \cite{Yu_2025_ICCV}
&88.19
&92.13
&93.49
&97.00
&0.661
&94.89
&\textbf{87.54}
&\underline{39.06} \\
Vista4D \cite{lin2026vista4d}
&\underline{88.81}
&\underline{92.28}
&93.22
&96.77
&\textbf{0.751}
&96.28
&86.26
&\textbf{51.25} \\
\textbf{EditStream Teacher}
&88.52
&\textbf{92.94}
&\underline{94.81}
&\underline{98.07}
&\underline{0.665}
&\underline{96.74}
&84.11
&38.84 \\

\midrule
\rowcolor{gray!15}
\multicolumn{9}{l}{\textbf{Streaming Models}} \\

\textbf{EditStream-AR} (ODE Init. + Self-Forcing Unrolling \cite{huang2026self})
&\underline{88.41}
&\underline{92.87}
&\textbf{94.74}
&\underline{97.58}
&\textbf{0.657}
&89.26
&81.41
&30.96 \\

\textbf{EditStream-AR (VMM Init. + Self-Forcing Unrolling)}
&88.23
&92.62
&94.67
&97.50
&\underline{0.633}
&\underline{93.16}
&\underline{81.68}
&\underline{33.00} \\

\textbf{EditStream-AR (VMM Init. + VMM-Forcing Unrolling)}
&\textbf{88.55}
&\textbf{92.90}
&\underline{94.70}
&\textbf{97.95}
&0.622
&\textbf{97.48}
&\textbf{84.03}
&\textbf{36.27} \\

\bottomrule
\end{tabular}
}
\end{table*}
\begin{figure*}[htbp]
    \centering
    \includegraphics[width=\textwidth]{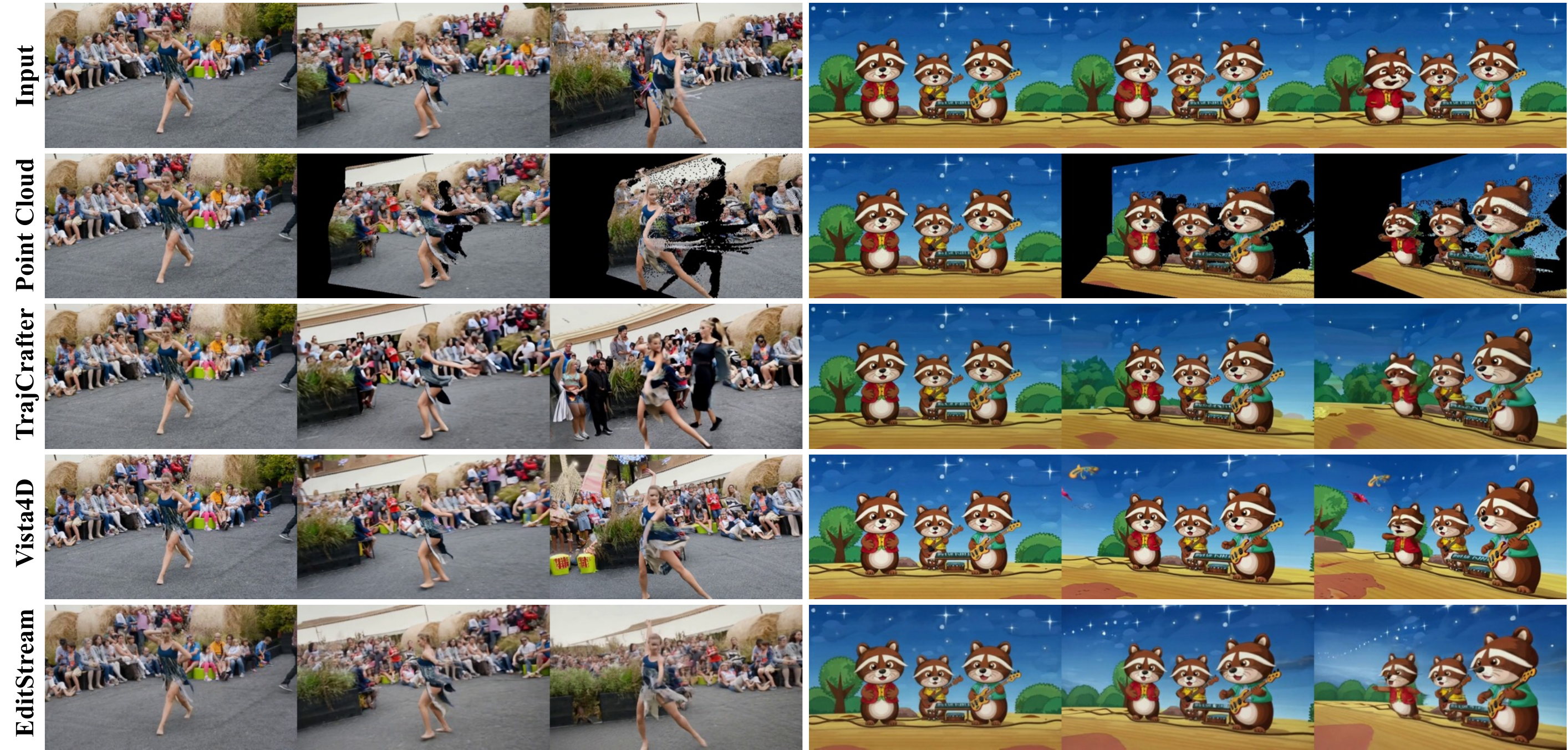}
    \caption{Camera Pose (ReShoot) Control Results on the collected V2VCameraPose Benchmark. TrajCrafter and Vista4D are both strong offline baselines. However, they occasionally suffer from over-hallucination, introducing implausible objects or environmental elements. EditStream generally produces milder distortions while avoiding unnecessary hallucinated content.}
    \label{fig:campose}
\end{figure*}
\textbf{Explicit Camera Pose Change (ReShoot).} 
We construct a video-to-video camera pose control benchmark (V2VCameraPose) comprising 472 test cases, obtained by applying 4 canonical camera trajectories (an orbiting \emph{loop}, a rightward \emph{orbit}, an upward \emph{pan}, and a \emph{zoom-in}) to 118 source clips. The source clips are drawn from both real-world and generative video domains: 78 clips from the DAVIS video dataset~\cite{davis}, and 20 clips each synthesized by the state-of-the-art video generation models Kling~v2.1 and Sora~2, ensuring the benchmark covers diverse content distributions and motion patterns beyond natural footage.

Following the camera-conditioned rendering pipeline of
TrajectoryCrafter~\cite{Yu_2025_ICCV}, for each source clip we estimate per-frame monocular depth and back-project the frames into a 3D point cloud, which is then re-rendered under each of the four target camera trajectories to produce a warped reference video together with a visibility mask marking the disoccluded regions that the model must hallucinate. The corresponding per-frame camera extrinsics/intrinsics are stored alongside each sample but not used in our inference. Each sample is further annotated with a dense caption produced by a vision-language captioner, which is used as the text prompt during evaluation.

Table~\ref{tab:camerapose} reports camera pose control results on the V2VCameraPose benchmark. The definition of the metrics can be found in the Appendix. The EditStream teacher and all causal student variants maintain strong subject and background consistency while achieving competitive temporal stability. Among the streaming variants, self-forcing achieves the highest perceptual quality in the observed regions, whereas VMM-forcing provides the best overall performance on dis-occluded region completion, temporal stability, and motion smoothness, indicating a better trade-off between fidelity and long-horizon consistency.

Qualitative comparisons in Figure \ref{fig:campose} further highlight the strengths and limitations of different approaches. TrajCrafter and Vista4D are both remarkably strong offline baselines and often produce highly detailed novel views. However, they occasionally suffer from over-hallucination, introducing implausible objects or environmental elements that are not supported by the input scene. Among them, EditStream generally produces milder distortions while avoiding unnecessary hallucinated content. Importantly, unlike existing offline methods, EditStream achieves these results in a low latency streaming setting, making it suitable for latency-sensitive interactive applications.

\begin{figure*}[htbp]
    \centering
    \includegraphics[width=\textwidth]{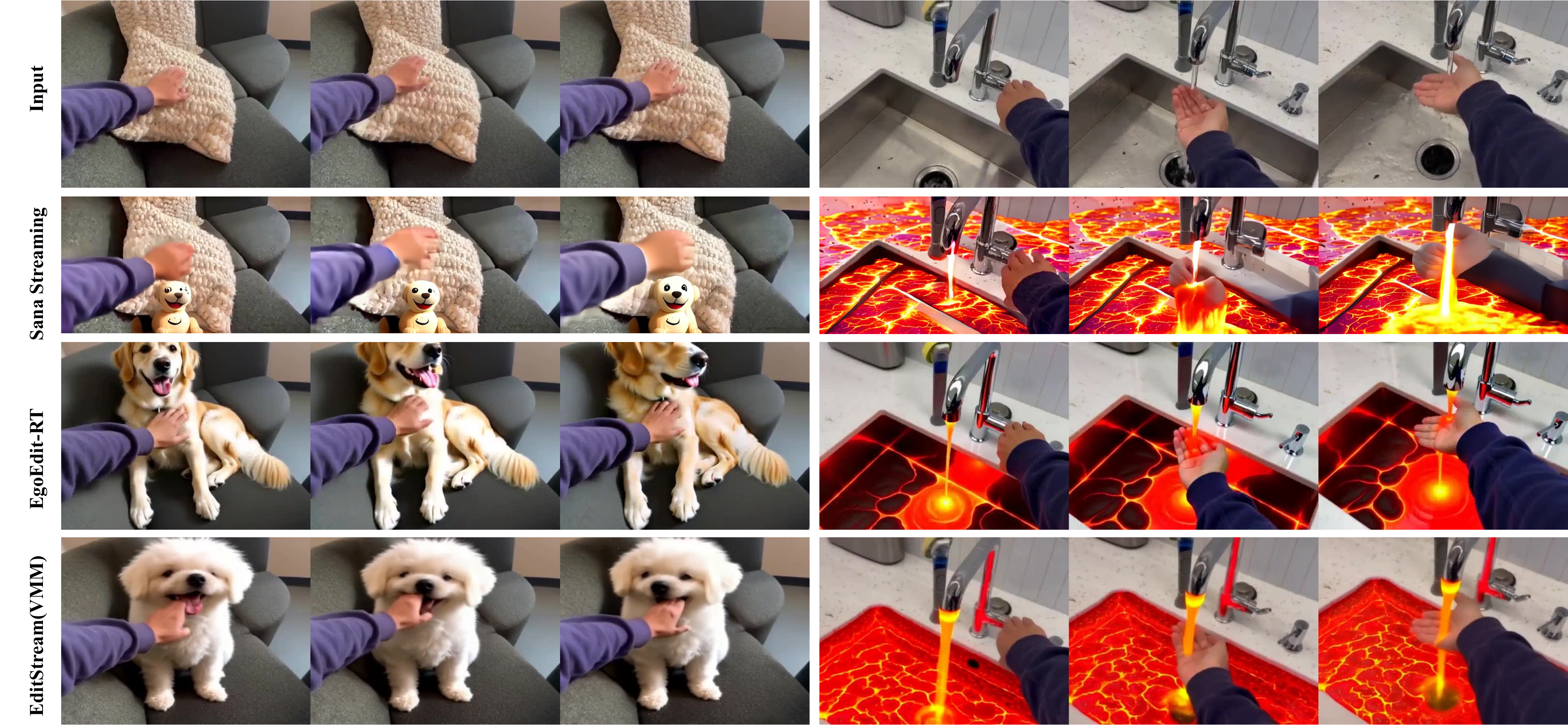}
    \caption{Egocentric Editing (Ego V2V) Results. EgoEdit tends to produce sharper outputs, but they often appear less realistic. It may partly reflect the sharpness and mode-seeking behavior associated with DMD-based training. In the dog editing example, EgoEdit generates overly pronounced fur textures and introduces unrealistic specular highlights on the dog's body. Similarly, in the lava editing example, the generated lava shows a CGI-like texture. EditStream with VMM produces outputs that are slightly less sharp but exhibit more faithful material properties, more realistic textures, and a more natural overall appearance.}
    \label{fig:egoedit}
\end{figure*}
\begin{figure*}[htbp]
    \centering
    \includegraphics[width=\textwidth]{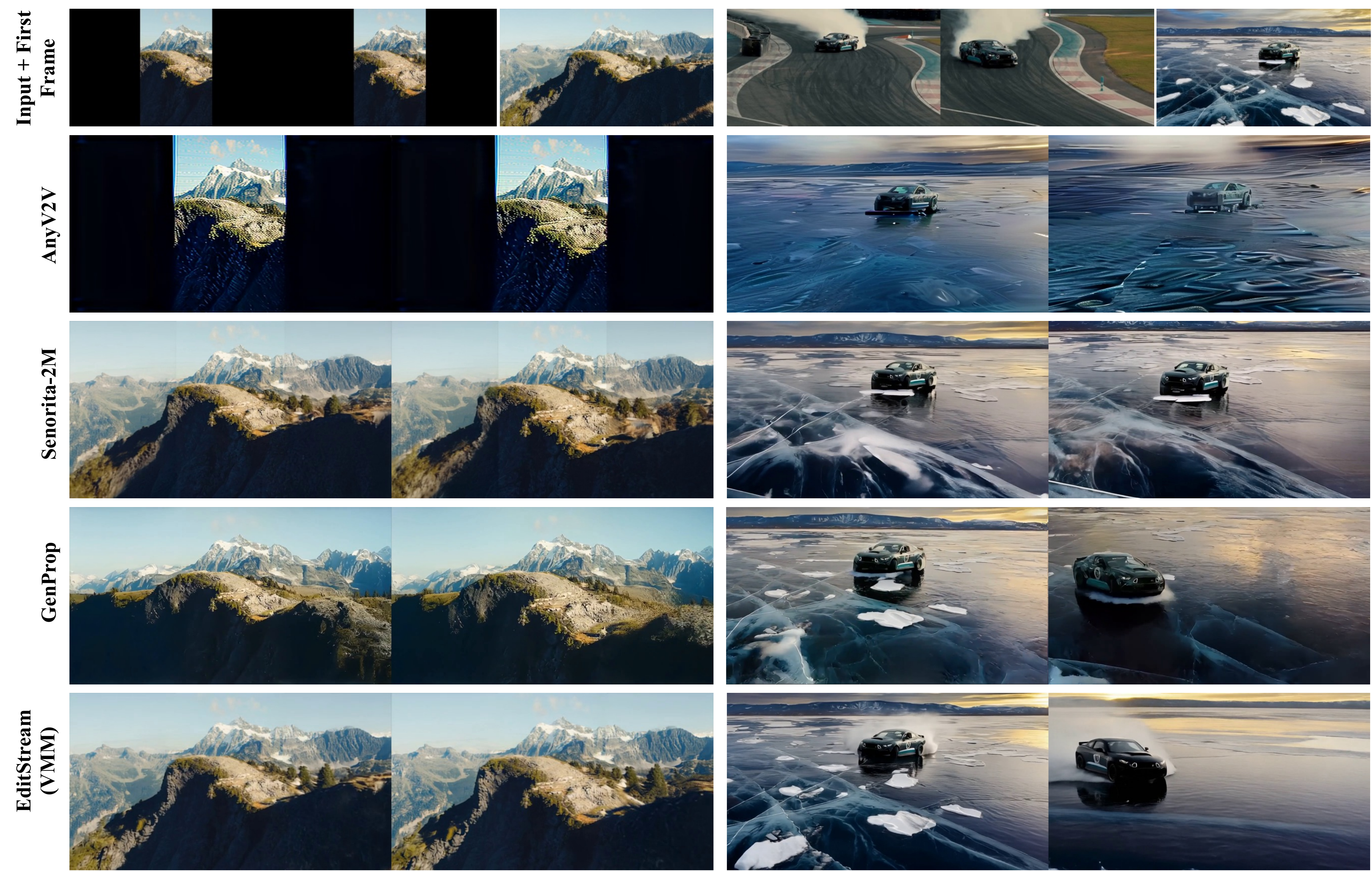}
    \caption{Editing Propagation Results. AnyV2V and Señorita-2M struggle in challenging scenarios. Although EditStream is not trained specifically for outpainting, it demonstrates strong generalization capability. EditStream results are more realistic and detailed than that of GenProp (e.g. it produces a more realistic effect behind the car). It may partly benefit from differences in the underlying foundation model. } 
    \label{fig:propresults}
\end{figure*}
\textbf{EgoCentric Video-to-Video (Ego-V2V) Editing}
We present qualitative comparisons on egocentric video editing using assets from EgoEditBench \cite{li2026egoedit} in Figure~\ref{fig:egoedit}. We compare our method against SANA Streaming and EgoEdit. Since EgoEdit has not been open-sourced, its results are reused from the official project website. Notably, all three methods are autoregressive and distilled to low-latency streaming inference.

Because our model is trained on the same egocentric editing dataset as EgoEdit, the editing quality is overall comparable. In contrast, SANA Streaming performs noticeably worse on egocentric-like editing, which is likely due to the differences in its training data.

A direct comparison of the distillation strategies used by EgoEdit and EditStream is challenging. Nevertheless, we observe consistent differences in their visual characteristics. EgoEdit tends to produce sharper outputs, but they often appear less realistic, which is consistent with the behavior of DMD-based training causing mode collapse issues. For example, in the dog editing example, EgoEdit generates overly pronounced fur textures and introduces unrealistic specular highlights on the dog's body, resulting in an artificial appearance. Similarly, in the lava editing example, the generated lava shows a CGI-like texture. In comparison, EditStream with VMM produces outputs that are slightly less sharp but exhibit more faithful material properties, more realistic textures, and a more natural overall appearance.

\textbf{Editing Propagation (EditProp) Results.}
Figure~\ref{fig:propresults} presents qualitative comparisons on video editing propagation from a single edited first frame. We did not conduct quantitative evaluation due to the lack of large-scale editing propagation benchmark. We compare EditStream with AnyV2V, Senorita-2M, and GenProp on the small-scale benchmark released by GenProp. Since GenProp has not been open-sourced, we reuse its results from the official project website.

We observe that AnyV2V and Señorita-2M struggle in challenging scenarios, particularly those involving large outpainting regions or significant background changes caused under fast camera or object motion. Although EditStream is not trained specifically for outpainting, it demonstrates strong generalization capability. In the outpainting case, EditStream produces seamless spatial extensions without noticeable boundary artifacts. Benefiting from a stronger foundation model, its generated content is also perceptually more realistic and detailed than that of GenProp (e.g. it produces a more realistic effect behind the car). Meanwhile, our model structure is simpler than the architecture design with external encoder and mask decoder in GenProp. More importantly, EditStream performs propagation in a low-latency interactive streaming manner, whereas existing approaches operate offline. To the best of our knowledge, this is the first streaming method evaluated on the GenProp benchmark.
\section{Ablation Studies}
\subsection{Data Ablation}
\begin{figure*}[htbp]
    \centering
    \includegraphics[width=\textwidth]{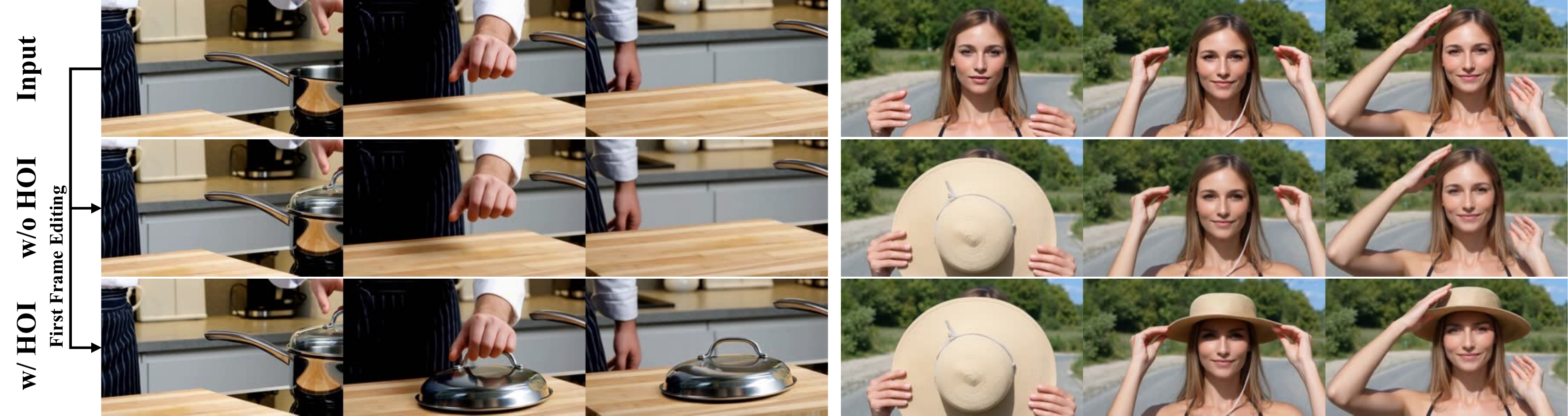}
    \caption{Editing Propagation Results for human-object interaction cases. Without the proposed HOI training data, the model struggles to establish stable physical correspondence between hands and edited objects. As a result, the generated object frequently disappears, fails to follow the intended manipulation, or exhibits severe geometric morphing during interaction. In contrast, incorporating HOI supervision significantly improves the realism and controllability of human--object interactions, making edited virtual objects behave naturally under user manipulation.}
    \label{fig:hoiresults}
\end{figure*}
\textbf{Effects of Human-Object Interaction (HOI) Data.}
To enable realistic interactions between users and newly generated virtual objects, we augment the training data with rendered human--object interaction (HOI) sequences during video-to-video editing and editing propagation training. These synthetic sequences provide explicit supervision for the physical relationship between human hand motions and edited objects, allowing the model to learn plausible object manipulation behaviors.

Figure~\ref{fig:hoiresults} demonstrates the effectiveness of this training data. For example, after generating a virtual pot lid in the first frame, the user can naturally grasp and lift it using hand motions. Likewise, a generated hat can be picked up and accurately placed onto the user's head. When the hand releases the object, the model is able to maintain temporally consistent object placement instead of forcing the object to remain attached to the hand. Such interaction behaviors emerge without any task-specific inference procedure.

Without the proposed HOI training data, the model struggles to establish stable physical correspondence between hands and edited objects. As a result, the generated object frequently disappears, fails to follow the intended manipulation, or exhibits severe geometric morphing and interpenetration artifacts during interaction. In contrast, incorporating HOI supervision significantly improves the realism and controllability of human--object interactions, making edited virtual objects behave naturally under user manipulation.

\subsection{Distillation Modules Ablation}
\textbf{Effects of Energy Annealing} Energy Annealing is essential to improve over-saturation issues caused by Classifier-free Guidance during both teacher inference and distillation. We conducted the ablation study on different energy annealing factors in the proposed EA strategy as shown in Table \ref{tab:color-quality}. In the table, we introduced multiple new metrics to better measure the color quality of different models. This is because standard perceptual video metrics (e.g.\ aesthetic/imaging quality scores from
VBench~\cite{huang2024vbench}) use a single scalar which makes it hard to diagnose and differentiate different color issues. 

The definition of the metrics can be found in the Appendix. All metrics are computed per frame on the generated RGB clip and averaged over the clip. We report the mean and standard deviation across all vbench videos. Let $I \in
[0,255]^{H\times W\times 3}$ denote a single RGB frame. We utilize the following metrics as shown in Table~\ref{tab:color-quality}. The detailed definition of these metrics is explained in the appendix. 

We compute the metrics above over all 946 prompts of the vbench \cite{huang2024vbench}
benchmark for three checkpoints from the
EA ablation: the \textbf{EA$=0.7$} baseline student model, \textbf{no-EA} (teacher
energy annealing disabled), and \textbf{EA$=0.2$} (energy-annealing target level
lowered from 0.7 to 0.2). 
\newsavebox{\CQTableBody}
\newsavebox{\CQTableCaption}
\newsavebox{\CQFigureCaption}

\newlength{\CQLeftWidth}
\newlength{\CQRightWidth}
\newlength{\CQTableCaptionHeight}
\newlength{\CQFigureCaptionHeight}
\newlength{\CQCaptionHeight}
\newlength{\CQBodyHeight}
\newlength{\CQCaptionBodyGap}

\setlength{\CQLeftWidth}{0.58\textwidth}
\setlength{\CQRightWidth}{0.38\textwidth}
\setlength{\CQCaptionBodyGap}{0.5em}

\sbox{\CQTableBody}{%
  \resizebox{\CQLeftWidth}{!}{%
    \begin{tabular}{lccc}
      \toprule
      Metric
      & EA$=0.7$ (ours)
      & No EA
      & EA$=0.2$ \\
      \midrule

      \rowcolor{gray!15}
      \multicolumn{4}{l}{\textbf{Color richness}} \\

      Colorfulness $M$ (Eq.~\ref{eq:colorfulness})
      & 60.03 $\pm$ 27.15
      & \textbf{75.78 $\pm$ 33.11}
      & 49.58 $\pm$ 21.68 \\

      Lab chroma $\bar C^*$ (Eq.~\ref{eq:chroma})
      & 21.86 $\pm$ 10.69
      & \textbf{26.24 $\pm$ 12.36}
      & 18.77 $\pm$ 9.21 \\

      HSV saturation $\bar S$
      & 0.485 $\pm$ 0.174
      & \textbf{0.563 $\pm$ 0.176}
      & 0.412 $\pm$ 0.157 \\

      Dynamic range $\mathrm{DR}$ (Eq.~\ref{eq:dynamic-range})
      & 0.882 $\pm$ 0.129
      & \textbf{0.933 $\pm$ 0.105}
      & 0.797 $\pm$ 0.138 \\

      \midrule

      \rowcolor{gray!15}
      \multicolumn{4}{l}{\textbf{Artifacts / Costs}} \\

      Shadow clipping (Eq.~\ref{eq:shaowclip})
      & 0.073 $\pm$ 0.104
      & 0.109 $\pm$ 0.109
      & \textbf{0.034 $\pm$ 0.078} \\

      Highlight clipping (Eq.~\ref{eq:hightlightclip})
      & 0.006 $\pm$ 0.025
      & 0.020 $\pm$ 0.046
      & \textbf{0.001 $\pm$ 0.004} \\

      Color flicker $\downarrow$ (Eq.~\ref{eq:flicker})
      & 3.16 $\pm$ 2.24
      & 3.77 $\pm$ 2.43
      & \textbf{2.83 $\pm$ 1.86} \\

      \bottomrule
    \end{tabular}%
  }%
}

\setlength{\CQBodyHeight}{%
  \dimexpr\ht\CQTableBody+\dp\CQTableBody\relax
}

\sbox{\CQTableCaption}{%
  \begin{minipage}[t]{\CQLeftWidth}
    \vspace{0pt}
    \captionof{table}{%
      \textbf{Color Quality}: Comparison of color-related metrics over 946 generations. The first four metrics are descriptive measures of color richness, for which bold marks the highest values; higher richness is not necessarily better. For the final three artifact metrics, bold marks the lowest values. 
    }
    \label{tab:color-quality}
  \end{minipage}%
}

\sbox{\CQFigureCaption}{%
  \begin{minipage}[t]{\CQRightWidth}
    \vspace{0pt}
    
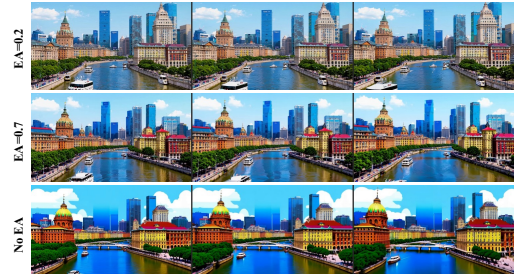
\captionof{figure}{%
      Visual comparison of sample generations under different EA settings. no-EA yields over-saturation. EA=0.2 yields a flatter and duller look. EA=0.7 balances the colorfulness.
    }
    \label{fig:color-example}
  \end{minipage}%
}

\setlength{\CQTableCaptionHeight}{%
  \dimexpr\ht\CQTableCaption+\dp\CQTableCaption\relax
}

\setlength{\CQFigureCaptionHeight}{%
  \dimexpr\ht\CQFigureCaption+\dp\CQFigureCaption\relax
}

\setlength{\CQCaptionHeight}{\CQTableCaptionHeight}

\ifdim\CQFigureCaptionHeight>\CQCaptionHeight
  \setlength{\CQCaptionHeight}{\CQFigureCaptionHeight}
\fi

\begin{figure*}[t]
  \centering

  \begin{minipage}[t]{\CQLeftWidth}
    \vspace{0pt}

    \begin{minipage}[t][\CQCaptionHeight][t]{\linewidth}
      \vspace{0pt}
      \usebox{\CQTableCaption}
      \vfill
    \end{minipage}

    \par\nointerlineskip
    \vspace*{\CQCaptionBodyGap}

    \noindent
    \usebox{\CQTableBody}
  \end{minipage}%
  \hfill
  \begin{minipage}[t]{\CQRightWidth}
    \vspace{0pt}

    \begin{minipage}[t][\CQCaptionHeight][t]{\linewidth}
      \vspace{0pt}
      \usebox{\CQFigureCaption}
      \vfill
    \end{minipage}

    \par\nointerlineskip
    \vspace*{\CQCaptionBodyGap}

    \noindent
    \makebox[\linewidth][c]{%
      \includegraphics[
        height=\CQBodyHeight,
        keepaspectratio
      ]{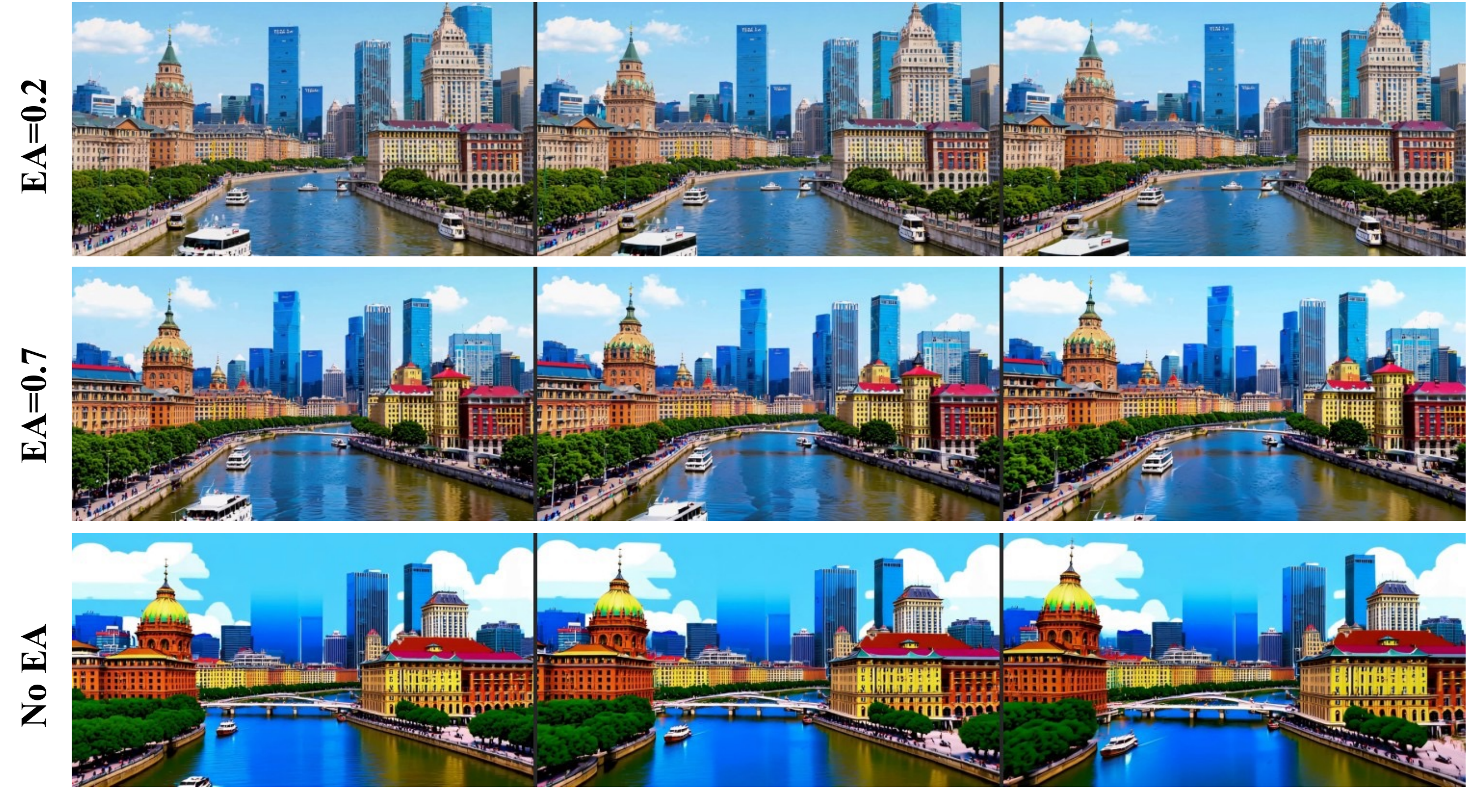}%
    }
  \end{minipage}

\end{figure*}

Critically, a model can maximize colorfulness and saturation simply by pushing pixel values toward over-saturation, which we show below is exactly what happens when the teacher's energy annealing (EA) is disabled. Table \ref{tab:color-quality} shows that disabling energy annealing (no-EA) yields the highest colorfulness. In the meanwhile, the variant has the highest shadow/highlight clipping and color flickering. This combination is the quantitative signature of over-saturation. In contrast, EA=0.2 has the lowest colorfulness range, yielding a flatter and duller look, though having the most temporally stable and least clipping. The baseline EA=0.7 sits between the two and balances the colorfulness and pixel stability. 

\begin{figure}[htbp]
  \centering
  \begin{minipage}[t]{0.49\linewidth}
    \centering
    \includegraphics[width=\linewidth]{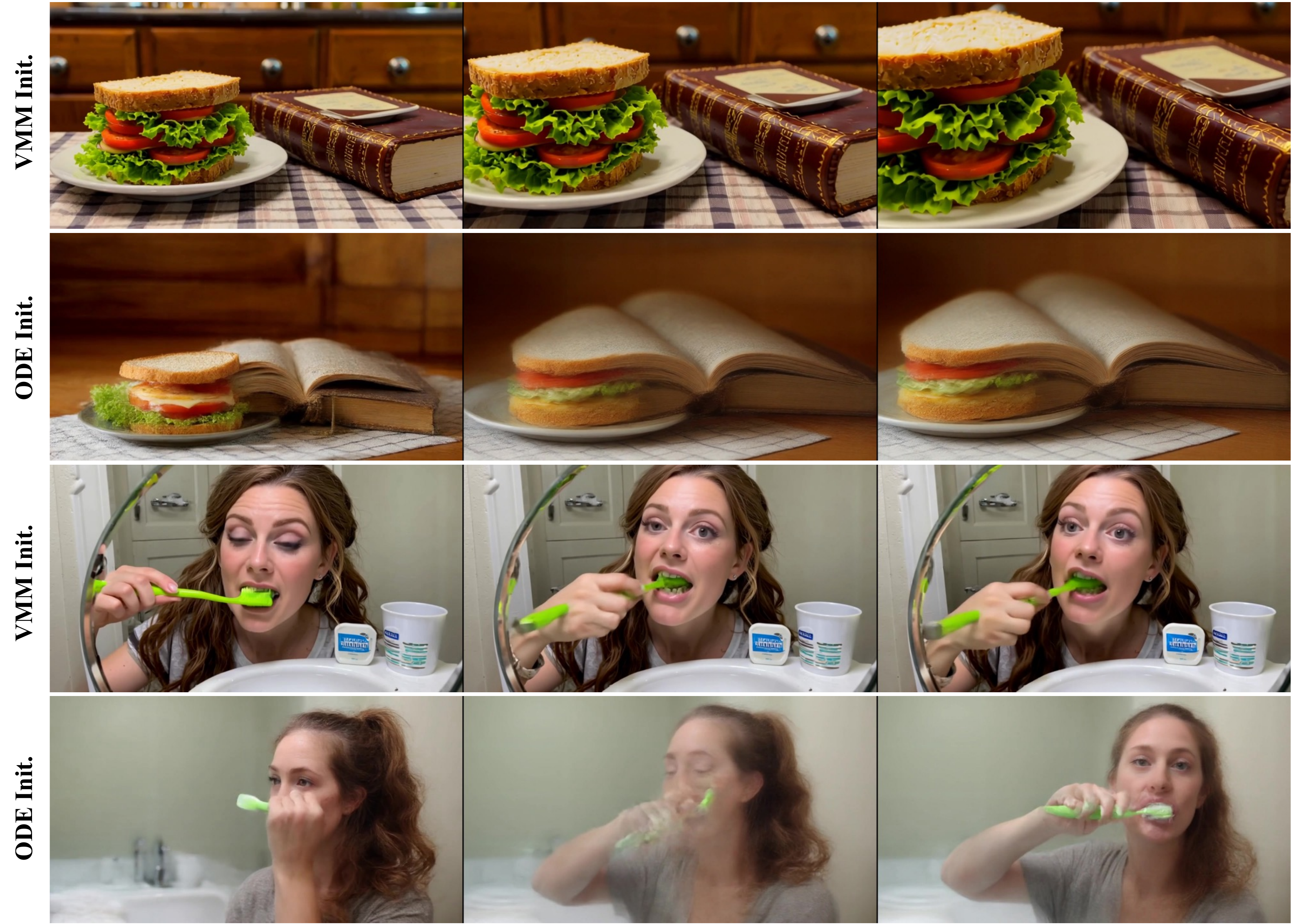}
    \caption*{(a) Quality of different initialization methods. The proposed VMM initialization provides a much stronger starting point.}
  \end{minipage}
 \begin{minipage}[t]{0.49\linewidth}
    \centering
    \includegraphics[width=\linewidth]{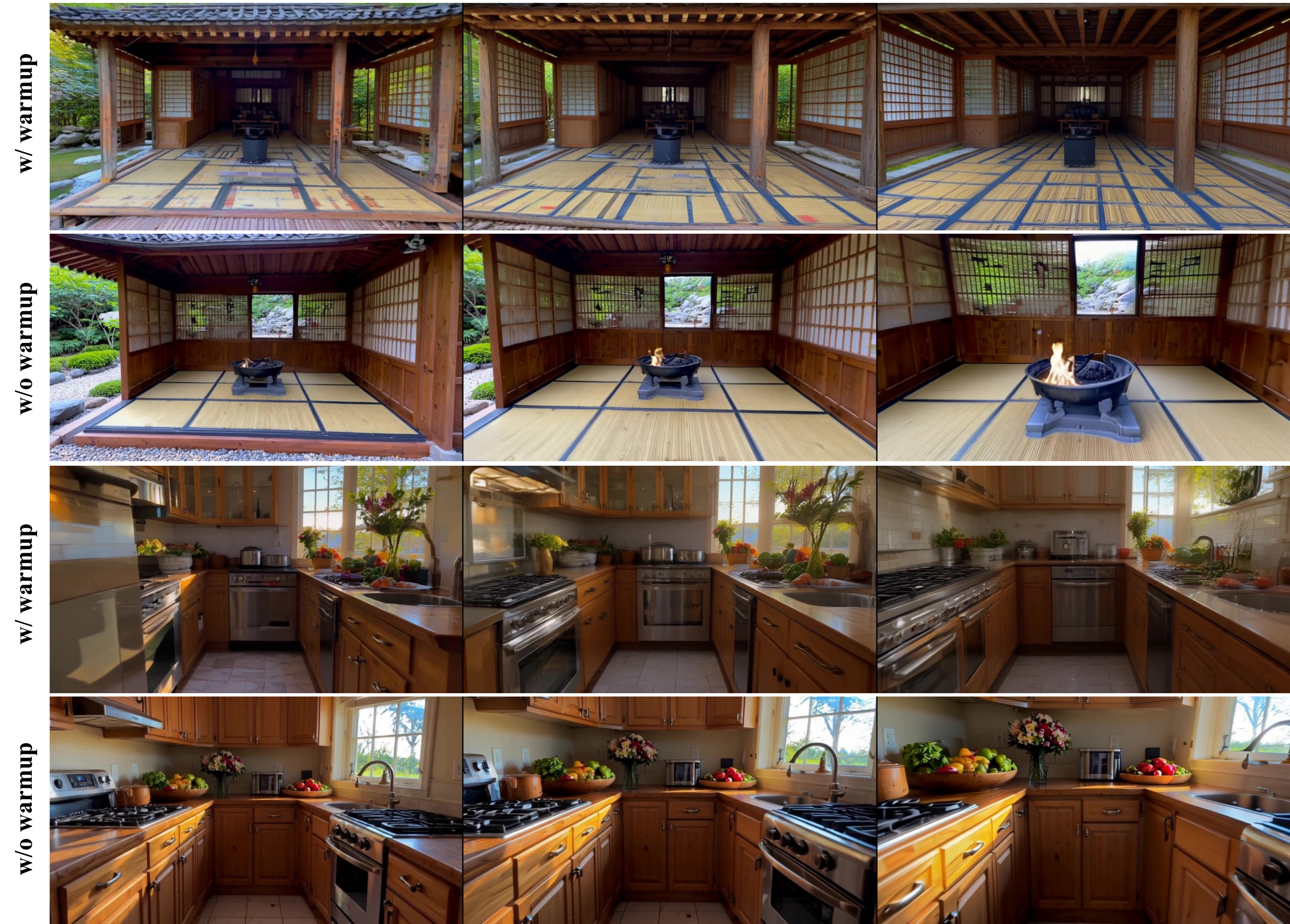}
    \caption*{(b) Importance of Warm-up. Eliminating warm-up introduces noticeable discontinuities and inconsistencies between blocks.}
  \end{minipage}\hfill
  \caption{Ablation study visualizations for warm-up existence and initialization methods.}
  \label{fig:abla_warmup_init}
\end{figure}

\begin{table*}[htbp]
\centering
\caption{\textbf{Ablation study} on the proposed training strategy. Higher is better for all metrics.}
\label{tab:ablation}
\setlength{\tabcolsep}{4.5pt}
\renewcommand{\arraystretch}{1.15}
\resizebox{0.98\textwidth}{!}{
\begin{tabular}{c|ccc|ccc|ccc|c|c}
\toprule

&
\multicolumn{6}{c|}{\textbf{Configuration}}
&
\multicolumn{5}{c}{\textbf{Results}}
\\
\toprule
&
\multicolumn{3}{c|}{\textbf{Warm-up}}
&
\multicolumn{3}{c|}{\textbf{Unrolling}}
&
\multicolumn{3}{c|}{\textbf{T2V}}
&
\textbf{V2V}
&
\textbf{RefV2V}
\\
\cmidrule(lr){2-4}
\cmidrule(lr){5-7}
\cmidrule(lr){8-10}

\textbf{ID}
&
\textbf{Method}
&
\textbf{Iter.}
&
\textbf{W/U Ratio}
&
\textbf{Method}
&
\textbf{Alt.}
&
\textbf{Commit}
&
\textbf{Quality}
&
\textbf{Semantic}
&
\textbf{Overall}
&
\textbf{VLM}
&
\textbf{Overall}
\\
\midrule

B1
& VMM Init. & 200 & 1:4 & VMM-forcing Unrolling & \ding{51} & \ding{51}
& \textbf{85.36} & 80.64 & \textbf{84.42} & \textbf{7.35} & 3.71
\\
\midrule

B2
& ODE Init. & 5K & 5:1 & Self-forcing Unrolling & N/A & N/A
& 84.09 & \textbf{81.95} & 83.67 & 7.01 & 3.18
\\

\midrule

E1
& ODE Init. & 5K & 1:0 & N/A & N/A & N/A
& 81.72 & 74.33 & 80.24 & 6.98 & 2.68
\\

E2
& VMM Init. & 200 & 1:0 & N/A & N/A & N/A
& 84.17 & 79.89 & 83.31 & 7.13 & 3.55
\\

E3
& No Warmup & 0 & 0 & VMM-forcing Unrolling & N/A & N/A
& 83.80 & 78.52 & 82.74 & 7.22 & 2.87
\\

E4
& VMM Init. & 200 & 1:4 & Self-forcing Unrolling & N/A & N/A
& \underline{85.03} & 80.52 & \underline{84.13} & \underline{7.25} & \underline{3.72}
\\

E5
& VMM Init. & 500 & 1:1 & VMM-forcing Unrolling & \ding{51} & \ding{51}
& 84.71 & \underline{80.85} & 83.94 & 7.24 & 3.67
\\

E6
& VMM Init. & 200 & 1:4 & VMM-forcing Unrolling & \ding{51} & \ding{55}
& 84.91 & 80.70 & 84.07 & 7.23 & 3.68
\\

E7
& VMM Init. & 200 & 1:4 & VMM-forcing Unrolling & \ding{55} & \ding{55}
& 84.73 & 80.43 & 83.87 & 7.27 & \textbf{3.73}
\\

\bottomrule
\end{tabular}}
\end{table*}
\textbf{Effect of Warm-up Initialization.}
We first compare different warm-up strategies without unrolling (E1 vs.~E2 in Table \ref{tab:ablation}). ODE initialization, adopted by previous autoregressive distillation methods such as Self-forcing, requires approximately 5K optimization iterations before reaching convergence, and its effectiveness depends on the scale of the precomputed ODE supervision dataset. In contrast, the proposed VMM initialization converges within only 200 iterations while already reaching a practically usable performance level. Compared with ODE initialization, VMM initialization improves the T2V overall score from \textbf{80.24} to \textbf{83.31}, the V2V score from \textbf{6.98} to \textbf{7.13}, and the RefV2V score from \textbf{2.68} to \textbf{3.55}. Consequently, the proposed initialization significantly reduces the optimization burden of the subsequent unrolling stage by providing a much stronger starting point as shown in Figure \ref{fig:abla_warmup_init}(a).

We further isolate the effect of initialization by comparing the original Self-forcing pipeline (B2) with the same Self-forcing unrolling trained from the proposed VMM initialization (E4). Replacing only the initialization consistently improves performance across all three tasks, increasing the T2V overall score from \textbf{83.67} to \textbf{84.13}, the V2V score from \textbf{7.01} to \textbf{7.25}, and the RefV2V score from \textbf{3.18} to \textbf{3.72}. Similar improvements can also be observed across the detailed metrics in Tables~\ref{tab:vbench_t2v}--\ref{tab:camerapose}. These results demonstrate that warm-up initialization is a critical component for autoregressive distillation, consistent with observations made in recent causal distillation methods \cite{zhu2026causalforcing}. We do not further compare with alternative initialization approaches such as Causal Forcing because they assume different teacher formulations.

We also investigate whether the warm-up stage can be removed by directly training with unrolling (E3 vs.~B1). Eliminating warm-up consistently degrades performance across all tasks, with the largest drop observed on generative tasks such as text-to-video and reference-guided editing. Qualitative results in Figure \ref{fig:abla_warmup_init}(b) further show that directly applying unrolling introduces noticeable discontinuities between autoregressive blocks, making it difficult for the model to learn stable self-convergence. These observations demonstrate that the warm-up stage is an essential prerequisite for efficient and stable autoregressive distillation.

\textbf{VMM-forcing versus Self-forcing Unrolling.}
We then compare different unrolling strategies under the same VMM initialization (B1 vs.~E4). The two methods achieve largely comparable performance on the conditional video editing tasks, suggesting that once a sufficiently strong initialization is available, the choice of unrolling strategy has only a limited impact. Nevertheless, VMM-forcing consistently performs better on text-to-video generation, improving the overall VBench score from \textbf{84.13} to \textbf{84.42}. As shown in Table~\ref{tab:vbench_t2v}, the improvements mainly originate from Dynamic Degree, Color, Human Action, and perceptual Quality, indicating that VMM-forcing better preserves temporal diversity and motion dynamics during autoregressive generation, making it closer to the teacher quality.

\textbf{Warm-up Schedule.}
We further study the allocation of the training budget between warm-up and unrolling by comparing different warm-up/unrolling ratios under the same total budget of 1K iterations (B1 vs.~E5). Specifically, we increase the warm-up stage from 200 to 500 iterations, corresponding to a warm-up/unrolling ratio of $1{:}1$ instead of $1{:}4$. A longer warm-up consistently improves semantic learning, increasing the T2V semantic score from \textbf{80.64} to \textbf{80.85}. We attribute this improvement to the additional first-block training during the VMM warm-up stage, which stabilizes semantic understanding and early-frame generation before autoregressive rollout. However, the gains remain relatively small and do not translate into better overall performance. Under the same 1K-iteration training budget, the original $1{:}4$ schedule consistently achieves better or comparable results across all three tasks, suggesting that allocating more optimization to unrolling is a more effective use of the training budget once a sufficiently strong initialization has been established.

\textbf{Committed Teacher Context and Alternating Distillation.}
Finally, we study two additional design choices introduced during the unrolling stage: (1) feeding the student's estimated first-block $\hat{x}_0$ to the teacher and auxiliary model as committed context, and (2) alternating between first-block and autoregressive distillation to continually preserve the warm-up capability. E6 removes the committed-context conditioning together with the corresponding cold-start masking, while E7 additionally removes the alternating first-block training.

Table~\ref{tab:ablation} shows that removing these components leads to a modest degradation on text-to-video generation compared with the full model (B1). Specifically, the T2V overall score decreases from \textbf{84.42} to \textbf{84.07} (E6) and \textbf{83.87} (E7), while the performance on video editing tasks remains largely unchanged, likely because these tasks are conditioned on substantially richer visual context. Overall, the quantitative results indicate that these design choices have a moderate impact on the final benchmark performance.

Their effect is more evident in the training dynamics. During the unrolling stage, removing committed-context conditioning consistently leads to higher and less stable unrolling losses, whereas the full model exhibits steadily decreasing training loss. In contrast, retaining alternating distillation alone without committed-context conditioning does not noticeably improve the optimization behavior in our experiments. These observations suggest that the proposed committed-context conditioning and alternating distillation primarily improve the stability of autoregressive optimization, while their influence on the final downstream performance is relatively modest.

\section{Limitations and Future Work}

EditStream provides a simple and effective framework for converting bidirectional video diffusion models into causal streaming models and distilling them for efficient inference. Its effectiveness has been validated across a broad range of video generation and editing tasks. Nevertheless, the current release has two main limitations:

\begin{itemize}
    \item \textbf{Long-context generation.} EditStream does not introduce specialized designs for long-context or long-memory modeling, nor does this report investigate strategies for mitigating error accumulation over long generation horizons. Consequently, the current model is primarily designed and evaluated for short-form video generation and editing.
    
    \item \textbf{Real-time high-resolution generation.} To support higher-quality generation at 720p resolution, we prioritize visual quality over further latency optimization. As a result, despite the use of few-step distillation, the released model is not yet optimized for real-time generation at this resolution.
\end{itemize}

These limitations are largely orthogonal to the core conversion and distillation framework developed in EditStream. Integrating complementary advances in long-context modeling and efficient high-resolution inference is therefore a promising direction for future work.
\section{Conclusion}

EditStream explores how diverse video generation and editing tasks can be unified within a single video DiT framework through dual-type conditioning. It further introduces a bidirectional-to-causal conversion framework based on our proposed Velocity Moment Matching (VMM) distillation algorithm, enabling streaming and interactive task inference. Experiments across text-to-video, video-to-video editing, reference-guided video editing, editing propagation, and camera pose control demonstrate that EditStream achieves high generation quality and inference efficiency. Meanwhile, the proposed conversion pipeline simplifies the distillation process. VMM warm-up provides consistent improvement over the original ODE-initialized Self-Forcing pipeline. Under the same VMM initialization, VMM-Forcing and Self-Forcing achieve comparable performance on conditional editing tasks, while VMM-Forcing better preserves motion and temporal diversity in text-to-video generation. Future work will explore more efficient inference and extend the framework to long-form video generation and editing.

\section{Acknowledgment}
We would like to thank Kai Zhang, Zhonghao Wang, and Yuchen Liu for their valuable contributions to the development of the VMM algorithm. We are grateful to Hailin Jin, Sylvain Paris, John Yang, Cynthia Lu, Jianming Zhang, Ashwin Ramesh and Yiru Shen for their insightful discussions, guidance, and continuous support throughout the development of the distillation framework.

\clearpage
\newpage
\bibliographystyle{ieeetr}
\bibliography{paper}

\appendix

\begin{center}
    {\LARGE\bfseries Appendix}
\end{center}
\vspace{1em}

\section{Implementation Details}
\label{sec:implementation}

\textbf{Teacher training.} We fully fine-tune all $5$B parameters of \texttt{Wan2.2-TI2V-5B} with FSDP (full sharding) on $8$ nodes $\times$ $8$ A100-80GB GPUs. Training operates on $704{\times}1280$ / $1280{\times}704$ resolution buckets at $81$ frames. We optimize a flow-matching ($v$-prediction) denoising loss with AdamW ($\text{lr}=2\times10^{-5}$, weight decay $0.01$, linear warmup followed by linear decay), effective batch is $192$, and maintain an EMA of the weights. $p_{first}=20\%$ and $p_{ps}=40\%$. 

\textbf{Student distillation.} The student is strictly block-causal. The distillation process follows the two stages introduced in the paper: VMM warm-up and VMM-forcing unrolling. The student, auxiliary, and teacher models in the distillation process are all initialized from the same EMA checkpoint of the teacher, where this checkpoint is obtained after teacher-forcing adaptation. The latent frames of the student are divided into chunks of three frames, and chunk-wise unrolling and KV caching are performed during both inference and training. Training adopts a 4-step timestep schedule, $t\in\{999,938,833,625\}$, which is aligned with the timestep schedule of the teacher's 4-step sampling after shifting. During training, the warm-up stage takes $200$ iterations, and the unrolling stage takes $800$ iterations. The total number of training iterations is $1000$. Training for longer does not lead to significant improvements in performance, while the training remains stable throughout. AdamW is likewise used as the optimizer, with an initial learning rate of $\text{lr}=2\times10^{-5}$, which is reduced to $4\times10^{-6}$ after entering the unrolling stage, with $\beta=(0.9,0.95)$ and a weight decay of $0.01$. The effective batch size is $128$. Training is conducted on $8$ nodes, each equipped with $8$ A100-80GB GPUs, for a total of $64$ GPUs. The update ratio between the auxiliary model and the student is $1{:}1$. In addition, the Energy-Annealing (EA) factor used to mitigate over-saturation is set to $0.7$.

\textbf{Self-forcing baseline.}
As a baseline we also distill a causal student with the original self-forcing recipe~\cite{huang2026self}. The
generator is warm-started from a causal ODE-regression checkpoint obtained via an offline Multi-task ODE-pair generation-then-regression procedure, rather than initializing directly from the bidirectional teacher's weights as in our recipe. The fake-score critic is updated $5\times$ more frequently than the generator ($\text{dfake\_gen\_update\_ratio}=5$, vs.\ $1$ in our recipe), with separate learning rates for generator and critic ($2\times10^{-6}$ / $4\times10^{-7}$), effective batch size $128$, on the same $8{\times}8$ A100-80GB setup. More details are listed in Table \ref{tab:impl-summary}.

\begin{table}[htbp]
\centering
\small
\caption{Implementation summary across the three training stages.}
\label{tab:impl-summary}
\begin{tabular}{@{}lccc@{}}
\toprule
 & Teacher & Student (ours) & Self-forcing baseline \\
\midrule
Init & Wan2.2-TI2V-5B pretrained & Teacher EMA weights & Causal ODE-regression ckpt \\
Causal / block size & No & Yes, 3 frames/block & Yes, 3 frames/block \\
Denoising steps & full schedule & 4 (\{999,938,833,625\}) & 4 (\{999,938,833,625\}) \\
Distillation loss & flow matching & VMM & DMD \\
Block unrolling & -- & yes (200-step warmup) & yes \\
Critic{:}generator ratio & -- & $1{:}1$ & $5{:}1$ \\
Learning rate & $2\times10^{-5}$ & $2\times10^{-5}\!\to\!4\times10^{-6}$ & $2\times10^{-6}$ (gen) / $4\times10^{-7}$ (critic) \\
Effective batch size & 192 & 128 & 128 \\
GPUs & $64\times$A100-80GB & $64\times$A100-80GB & $64\times$A100-80GB \\
\bottomrule
\end{tabular}
\end{table}

\section{New Evaluation Metrics}
\subsection{Camera Pose Control Metrics}
Camera pose change is achieved by warping the frames captured from the original camera to a new camera pose through reprojection. Therefore, the existing frame pixels form the \emph{valid} region, while regions that were previously occluded or did not exist in the original frames and thus need to be generated form the black-pixel \emph{hole} region. We provide a binary mask, $M_t\in{0,1}^{H\times W}$, where the mask value is $1$ in the hole region. VBench's overall evaluation metrics typically average over both the valid and hole regions, causing the results to be dominated by the valid region while diluting the contribution of the generated region. Therefore, we introduce four new mask-based metrics.

\emph{Hole Complete.} measures whether the hole region has been completely filled or remains
black as before:
\begin{equation}
\text{HoleComplete} = 100 \cdot \frac{1}{|\mathcal{T}_h|}\sum_{t\in\mathcal{T}_h}
\frac{\big|\{p : M_t(p)=1,\ \bar{Y}_t(p) \ge \tau\}\big|}
{\big|\{p : M_t(p)=1\}\big|}.
\end{equation}
where $\bar{Y}_t \in [0,255]$ denotes the grayscale intensity, $\tau=10$ is a near-black
threshold, and $\mathcal{T}_h = \{t : |\{p:M_t(p)=1\}| > 0\}$ is the set of frames that
actually contain hole pixels (frames with an all-valid mask are excluded from the temporal
average rather than contributing a value). The metric reaches $100\%$ if and only if all hole
pixels throughout the video clip are rendered with new content.

\emph{Hole Temp. Stab.} mainly measures whether the temporal consistency of the generated
content is on par with that of real video. For each consecutive frame pair $(t-1,t)$ we run
Farneback dense optical flow~\cite{farneback2003two} on the grayscale frames, forward-warp
frame $t-1$ into frame $t$, and take the per-pixel absolute warp error. $e_t^{\text{hole}}$ and
$e_t^{\text{valid}}$ are this error averaged over the pixels that are hole (resp.\ valid) in
\emph{both} frames $t-1$ and $t$ (no explicit forward--backward flow-consistency check is
performed; restricting to pixels whose category agrees across the pair is what keeps the
comparison meaningful):
\begin{equation}
\text{HoleTempStab} = 100 \cdot \frac{1}{|\mathcal{T}_s|}\sum_{t\in\mathcal{T}_s}
\min\!\Big(1,\ \frac{e^{\text{valid}}_t}{e^{\text{hole}}_t}\Big).
\end{equation}
where $\mathcal{T}_s = \{t\in\{1,\dots,T-1\} : e_t^{\text{hole}} > \epsilon\}$ with
$\epsilon=10^{-6}$; frame pairs with an empty hole/valid intersection, or with degenerate
(near-zero) hole-region flow error, are excluded from the average rather than being
epsilon-smoothed. A score of $100$ indicates that the hole region exhibits no more temporal
flickering than the genuinely observed region of the same video.

\emph{Valid SSIM} mainly measures the preservation quality of the valid region, i.e., whether
the appearance of the original video region has been altered. We compute the SSIM between the
generated video $Y_t$ and the point-cloud-rendered reference $R_t$, restricted to the valid
region:
\begin{equation}
\text{ValidSSIM} = \frac{1}{T}\sum_t \text{SSIM}\big(Y_t, R_t \,\big|\, M_t=0\big).
\end{equation}

\emph{Gated Hole Fill Qual.} This metric mainly measures whether the generated content in the
hole region is sufficiently plausible. The metric is ``gated'' because we also require the
model to preserve the valid region; otherwise, methods that regenerate the entire video may
instead achieve high scores, making it difficult to properly evaluate their hole-filling
quality. For each frame we take the largest 8-connected hole component (discarding components
smaller than $32\times32$\,px or $1024$\,px$^2$ in area, since BRISQUE is unreliable on tiny
crops), crop its tight bounding box (no padding, no resizing), and require at least $90\%$ of
that component's pixels to already be non-black ($\bar{Y}_t\ge\tau$); frames failing this fill
gate are assigned a hole-fill score of $0$ directly, since BRISQUE does not reliably penalize
partially-black crops. Otherwise we apply BRISQUE~\cite{mittal2012no}, a no-reference quality
assessment model based on natural-scene statistics, to the crop and map the raw score
$s_{\text{BRISQUE}}$ (lower is better) to a $0$--$100$ scale:
\begin{equation}
\text{HoleFillQual}_{\text{BRISQUE}} = 100\cdot\mathrm{clip}\!\Big(1-\frac{s_{\text{BRISQUE}}}{s_{\max}},\,0,\,1\Big),
\end{equation}
with $s_{\max}=120$ calibrated so that real photographic content in our data lands in
$s_{\text{BRISQUE}}\!\approx\![35,60]$; scores at or above $s_{\max}$ (including degenerate
``zero-variance'' crops, e.g.\ still-flat/black patches, for which BRISQUE is undefined) are
floored to $0$. Per-frame scores are averaged over frames, then multiplied by Valid SSIM:
\begin{equation}
\text{GatedHoleFillQual} = \text{HoleFillQual}_{\text{BRISQUE}} \times \text{ValidSSIM}.
\end{equation}
We do not use MUSIQ~\cite{ke2021musiq} here because it cannot effectively penalize regions
that remain ungenerated, such as black regions. Our experiments show that when the regions to
be generated remain entirely black, this learned metric cannot reliably distinguish between
real images and images containing black holes. In contrast, BRISQUE can effectively
distinguish between these cases.

\subsection{Color Quality of Videos}

\emph{Colorfulness.} We use the Hasler--S\"usstrunk metric~\cite{hasler2003measuring} to measure colorfulness. With opponent channels $rg = R - G$ and $yb = \tfrac{1}{2}(R+G) - B$,
\begin{equation}
M \;=\; \sqrt{\sigma_{rg}^2 + \sigma_{yb}^2} \;+\; 0.3\sqrt{\mu_{rg}^2 + \mu_{yb}^2},
\label{eq:colorfulness}
\end{equation}
where $\mu_{(\cdot)}, \sigma_{(\cdot)}$ are the spatial mean and standard
deviation of each opponent channel over the frame. Larger $M$ means a more saturated frame and smaller $M$ means a flatter frame.

\emph{CIELAB chroma and HSV saturation.}
We convert $I$ to
CIELAB and compute per-pixel chroma
\begin{equation}
C^*(x) \;=\; \sqrt{a^*(x)^2 + b^*(x)^2}, \qquad \bar C^* = \operatorname{mean}_x\, C^*(x),
\label{eq:chroma}
\end{equation}
and report the mean HSV saturation $\bar S = \operatorname{mean}_x S(x)$.

\emph{Luminance dynamic range and exposure clipping.}
Let $Y(x)\in[0,1]$ denote the Rec.~709 luma \cite{recommendation2015709} of pixel $x$. We report a
percentile-based, exposure-invariant dynamic-range score:
\begin{equation}
\mathrm{DR}
=
\frac{P_{95}(Y)-P_{5}(Y)}{P_{95}(Y)+P_{5}(Y)+\epsilon}.
\label{eq:dynamic-range}
\end{equation}
We also measure the fraction of pixels in shadow and highlight clipping regions:
\begin{equation}
\mathrm{ShadowClip}
=
\frac{1}{N}\sum_x \mathbbm{1}\{Y(x)<\tau_{\text{dark}}\},
\label{eq:shaowclip}
\end{equation}

\begin{equation}
\mathrm{HighlightClip}
=
\frac{1}{N}\sum_x \mathbbm{1}\{Y(x)>\tau_{\text{bright}}\},
\label{eq:hightlightclip}
\end{equation}
where $\tau_{\text{dark}}=0.02$ and $\tau_{\text{bright}}=0.98$.

\emph{Temporal color flicker.}
It measures color flickering across frames. We first estimate dense optical flow using the Farneback method~\cite{farneback2003two}. Given two consecutive frames $I_t$ and $I_{t+1}$, we compute the optical flow $F_{t+1\to t}$ from $I_{t+1}$ to $I_t$, and then perform backward warping using $F_{t+1\to t}$. We then compute the mean CIEDE2000 color difference~\cite{sharma2005ciede2000} between the warped frame and $I_t$ in the CIELAB color space:

\begin{equation}
\mathrm{ColorFlicker}
=
\frac{1}{T-1}
\sum_{t=1}^{T-1}
\operatorname{mean}_x
\Delta E_{00}
\bigl(
I_t(x),
\mathcal{W}(I_{t+1}, F_{t+1\rightarrow t})(x)
\bigr),
\label{eq:flicker}
\end{equation}
where $\mathcal{W}(I,F)$ denotes backward warping of image $I$ using the estimated optical flow field $F$, $T$ is the total number of frames, and $\Delta E_{00}$ denotes the CIEDE2000 color difference. And the lower of this value, the better.

\section{Distillation Pipeline Comparison}
\subsection{A Unified View of Related Distillation Objectives}
\label{sec:unified_distillation}

VMM defines a different training objective from related few-step
distillation methods, including Distribution Matching Distillation
(DMD) \cite{yin2024onestep, yin2024improved}, Consistency Distillation (CD) \cite{song2023consistency, zhao2026causalforcingp, kim2024consistencytrajectory}, and MeanFlow \cite{geng2025meanflow}. These methods differ
mainly in two aspects: the state on which supervision is applied and the
quantity being matched.

\textbf{Relation to Distribution Matching Distillation (DMD).}
The distillation scheme of VMM is closely related to
DMD~\cite{yin2024onestep}, sharing a similar teacher--fake residual
learning scheme. However, their matching objectives and matching states
are different. In DMD, after obtaining an estimated clean endpoint
\(\hat{\mathbf{x}}_0^S\) from the student, which can be induced from
\(v_t^S\), an independent DMD noise level r is sampled and the
endpoint is re-noised as
\begin{equation}
\mathbf{y}_r^S
=
(1-\sigma_r)\hat{\mathbf{x}}_0^S
+
\sigma_r\boldsymbol{\epsilon},
\qquad
\boldsymbol{\epsilon}
\sim
\mathcal{N}(\mathbf{0},\mathbf{I}).
\label{eq:dmd_renoising}
\end{equation}
The teacher model and a trainable fake-score model, which plays a role
similar to the auxiliary model in VMM, are evaluated at the same noisy
state \((\mathbf{y}_r^S,r)\). Their score difference provides the
gradient for matching the diffused endpoint marginals of the student
and teacher.

Here, \(r\) is a randomly sampled noise scale used to examine the
endpoint distribution. It is not the destination of a transition from
the source timestep \(t\). DMD therefore provides a multi-scale
distributional correction to the student's endpoint distribution, but
does not explicitly constrain the intermediate states or ODE trajectory
used to produce that endpoint.

VMM retains a teacher--auxiliary residual structure, but evaluates the teacher and auxiliary velocities at an intermediate state produced
by the few-step student sampler. Specifically, \(s<t\) is sampled within
the current inference interval, and both models are evaluated at the same student-induced state \((\mathbf{x}_s^S,s)\). VMM therefore applies a velocity-moment correction at states reached by the student. This is intended to encourage teacher-consistent local dynamics during few-step inference.

\textbf{Relation to Consistency Distillation, CTM, and MeanFlow.}
Consistency Distillation~\cite{song2023consistency,lu2025simplifying}
learns a clean-endpoint predictor by enforcing consistent predictions between states on the same teacher ODE trajectory. In discrete CD, the
lower-time state is obtained from the higher-time state using a teacher ODE solver, whereas continuous-time CD takes the infinitesimal limit of this consistency relation and evaluates the resulting trajectory derivative using a Jacobian--vector product (JVP). The learned consistency function represents a global mapping from an arbitrary timestep \(t\) to the clean endpoint.

Consistency Trajectory Models (CTM)~\cite{kim2024consistencytrajectory} generalize this fixed-endpoint parameterization by directly learning an anytime-to-anytime flow map\(G_\theta(\mathbf{x}_t,t,s)\approx\mathbf{x}_s\) for \(s<t\). MeanFlow~\cite{geng2025meanflow} instead represents arbitrary finite-time transport through an average-velocity field. In the teacher-distillation adaptation considered here, MeanFlow predicts the average teacher velocity over the finite interval from \(t\) to \(s\), which equivalently determines the teacher state reached at timestep
\(s\).\footnote{
The original MeanFlow is teacher-free and is trained from scratch. Here, we refer specifically to a teacher-distillation adaptation in which the finite-interval average-velocity target is obtained from a pretrained teacher trajectory. }
Thus, CD learns a \(t\!\rightarrow\!0\) endpoint map, whereas CTM and
teacher-distilled MeanFlow parameterize arbitrary \(t\!\rightarrow\!s\) finite-time transport through a state map and an
average-velocity field, respectively.

VMM also uses a pair of coupled timesteps \((t,s)\), but the role of the destination state is different. CD, CTM, and teacher-distilled
MeanFlow construct supervision from a teacher or reference trajectory. In contrast, VMM first produces \(\mathbf{x}_s^S\) using the student's
own transition and then evaluates the teacher--auxiliary velocity residual at this state. Therefore, in CTM and teacher-distilled
MeanFlow, the reference finite-time transition determines the supervision target; in VMM, the student destination \(\mathbf{x}_s^S\) instead serves as the query state at which the velocity correction is evaluated.

\textbf{Unified Fixed-Point View.}
Let
\begin{equation}
\mathbf{x}_s^T
=
\Phi_T^{t\rightarrow s}(\mathbf{x}_t),
\qquad
\mathbf{x}_s^S
=
\Psi_S^{t\rightarrow s}(\mathbf{x}_t)
\label{eq:teacher_student_flow_maps}
\end{equation}
denote the destinations produced by the teacher and student respectively. The teacher's average velocity within the interval is
\begin{equation}
u_T^{t\rightarrow s}(\mathbf{x}_t)
=
\frac{
\Phi_T^{t\rightarrow s}(\mathbf{x}_t)-\mathbf{x}_t
}{
\sigma_s-\sigma_t
}.
\label{eq:teacher_average_velocity}
\end{equation}
Equivalently,
\begin{equation}
\mathbf{x}_s^T = \Phi_T^{t\rightarrow s}(\mathbf{x}_t)
=
\mathbf{x}_t
+
(\sigma_s-\sigma_t)
u_T^{t\rightarrow s}(\mathbf{x}_t).
\end{equation}

For DMD, let \(\mathbf{x}_0^M\sim p_M\), where
\(M\in\{S,T\}\), denote a student or teacher endpoint sample. We define
its diffused marginal at noise level \(r\) as
\begin{equation}
p_{M,r}
:=
\operatorname{Law}
\left(
(1-\sigma_r)\mathbf{x}_0^M
+
\sigma_r\boldsymbol{\epsilon}
\right | \mathcal{C} ),
\qquad
\boldsymbol{\epsilon}
\sim
\mathcal{N}(\mathbf{0},\mathbf{I}).
\label{eq:diffused_marginal}
\end{equation}

For VMM, define the conditional first moment of the student transition
velocity as
\begin{equation}
m_S(\mathbf{x},s,\mathcal{C})
=
\mathbb{E}
\left[
\bar v_S^{t\rightarrow s}
\mid
\mathbf{x}_s^S=\mathbf{x},
s,
\mathcal{C}
\right].
\label{eq:vmm_conditional_moment}
\end{equation}

The ideal fixed points of the four objectives can then be summarized as
\begin{equation}
\begin{aligned}
\mathrm{CD:}&\qquad
f_S(\mathbf{x}_t,t,\mathcal{C})
\approx
\Phi_T^{t\rightarrow0}(\mathbf{x}_t,\mathcal{C}),
\\
\mathrm{MeanFlow:}&\qquad
u_S(\mathbf{x}_t,t,s,\mathcal{C})
\approx
u_T^{t\rightarrow s}(\mathbf{x}_t),
\\
\mathrm{DMD:}&\qquad
p_{S,r}
\approx
p_{T,r},
\\
\text{\bfseries VMM (ours):}&\qquad
m_S(\mathbf{x}_s^S,s,\mathcal{C})
\approx
v_T(\mathbf{x}_s^S,s,\mathcal{C}).
\end{aligned}
\label{eq:unified_fixed_points}
\end{equation}

The equations above characterize the ideal solutions of the respective objectives, while their practical training and optimization procedures are often more involved to ensure stable training. Broadly, CD and MeanFlow perform sample-wise flow-map matching, using teacher trajectories to specify target states. DMD performs distribution-wise endpoint matching without requiring sample-wise regression. VMM, in contrast, matches a conditional velocity moment at an intermediate state generated by the student.

Because CD and MeanFlow provide paired supervision from teacher trajectories, they generally offer stable training and help preserve the modes and diversity of the teacher distribution. However, since their objectives regress either a clean endpoint or a finite-time teacher flow map, few-step distillation or limited student capacity may lead to smoother outputs.

In contrast, DMD directly optimizes the distribution of student-generated outputs through a teacher--fake score correction. Since the objective tends to concentrate updates in high-density regions of the teacher distribution, it often produces sharper and perceptually higher-quality samples, while also carrying a higher risk of mode dropping or mode collapse~\cite{yin2024onestep,yin2024improved}. In autoregressive video generation, this behavior may also lead to low-motion solutions.

VMM falls somewhere in between. Similar to DMD, it applies a teacher--auxiliary residual correction at student-generated states, which can improve local generation quality and recover fine details. At the same time, because the matching state is associated with a specific t→s student inference transition, VMM also constrains the local sampling dynamics. VMM is therefore expected to strike a balance between perceptual quality and teacher-consistent few-step trajectories. Nevertheless, its final generation quality still depends on factors such as timestep scheduling during optimization and the stability of the auxiliary model.

\subsection{VMM initialization versus Others}

\textbf{Relation to Existing Causal Few-Step Initializations.}
Existing autoregressive diffusion distillation methods commonly construct a causal few-step student before self-forcing refinement. These initialization methods differ mainly in how the denoising states are constructed and whether the student is trained to regress a fixed teacher flow map or to correct its own transition.

\emph{ODE-pair initialization.}
CausVid and Self-Forcing~\cite{yin2025slow,huang2026self} initialize the causal student using ODE pairs generated by a bidirectional teacher. Intermediate states are precomputed from the teacher trajectory, and the student is trained to regress the corresponding clean endpoint. Different blocks can be selected at different stored ODE timesteps similar to Diffusion Forcing. Direct endpoint regression can be underdetermined for the causal student. Causal Forcing~\cite{zhu2026causalforcing} addresses the issues by first training an autoregressive diffusion teacher with
teacher forcing, and then generating causal ODE pairs under the same clean-prefix context used by the student. This produces a better-aligned flow-map target, but requires a separate causal-teacher training stage and offline generation.

\emph{Causal consistency initialization.}
Causal Forcing++~\cite{zhao2026causalforcingp} replaces full ODE-pair regression with causal Consistency Distillation. Causal ODE regression and causal CD learn the same autoregressive
teacher flow map, but approach it differently. ODE initialization
regresses from an intermediate state directly to the clean endpoint, whereas causal CD propagates the endpoint map through local consistency between adjacent timesteps. The latter avoids precomputing complete teacher trajectories and reduces the optimization gap in each training pair.

Causal-rCM~\cite{zheng2026causalrcm} follows the same general
teacher-forcing consistency route. It first converts the bidirectional model into a multi-step causal diffusion teacher, then initializes the few-step student using TF-dCM or continuous-time variants such as TF-sCM and TF-MeanFlow. It requires additional JVP-compatible training infrastructure.

\emph{VMM warm-up.}
The proposed VMM warm-up does not precompute teacher's ODE pairs, train a separate causal teacher, regress the student to a paired teacher destination, or use JVP-based consistency training. It indeed requires a teacher-forcing adaptation but the teacher itself can be retained bidirectional and the adaptation is efficient. Also, the same VMM objective can be retained when moving from warm-up to unrolling. 

Note that the bidirectional teacher and auxiliary model can use the clean suffix, while the causal student cannot. Consequently, VMM avoids direct regression to a future-dependent bidirectional endpoint, so the frame-level injectivity requirement of ODE-pair regression~\cite{zhu2026causalforcing} does not directly apply. However, the teacher--auxiliary residual can still contain future information, and the causal student effectively learns its projection onto the information available from the causal prefix. But to further address the teacher-forcing exposure gap, a subsequent unrolling stage is still necessary and preferred.

\subsection{VMM-forcing Unrolling versus Self-forcing Unrolling}
Our unrolling procedure is inspired by Self-Forcing
~\cite{huang2026self}, and both methods address exposure bias by generating the autoregressive history with the student itself and maintaining that history through a causal KV cache. However, they differ in both the within-block rollout strategy and the distillation objective.

Original Self-Forcing randomly samples a denoising exit index
\(q\in\{1,\ldots,J\}\). For every temporal block, it runs the diffusion trajectory only from the initial noise level \(\tau_J\) to \(\tau_q\), predicts a clean endpoint from that exit state, and commits this prediction to the KV cache:
\begin{equation}
\mathbf{x}_{\tau_J}^{k}
\longrightarrow
\cdots
\longrightarrow
\mathbf{x}_{\tau_q}^{k}
\longrightarrow
\widehat{\mathbf{x}}_{0}^{k,q}
\longrightarrow
\operatorname{KV}.
\end{equation}
The temporal autoregressive rollout covers all blocks, but the within-block denoising depth is truncated. Sampling
different exit indices across iterations allows different denoising steps to receive gradients while reducing training cost. A holistic distribution-matching objective, such as DMD, SiD, or a GAN loss, is then applied to the generated video.

In contrast, our VMM unrolling completes the deployed few-step trajectory for every block before committing its output:
\begin{equation}
\mathbf{x}_{\tau_J}^{k}
\longrightarrow
\mathbf{x}_{\tau_{J-1}}^{k}
\longrightarrow
\cdots
\longrightarrow
\mathbf{x}_{0}^{k}
\longrightarrow
\operatorname{KV},
\end{equation}
while VMM is evaluated on a selected local transition
\(t\rightarrow s\) along the trajectory. Consequently, the history used by every later block is produced by the same complete sampler used at inference, rather than by an endpoint prediction obtained from a randomly truncated denoising trajectory.

The two approaches also optimize different quantities according to the difference of DMD and VMM. Self-Forcing with DMD directly matches the distribution of complete student-generated videos and is free to learn a trajectory-independent generation shortcut. VMM unrolling instead applies a teacher--auxiliary velocity residual at intermediate states produced by the student's current few-step trajectory, which makes the training more stable. 

Importantly, once a sufficiently strong causal few-step initialization is available, both self-forcing unrolling and
VMM unrolling are valid post-training candidates. DMD provides a
direct endpoint-distribution correction and may be preferable when the main goal is perceptual quality or sharpness, whereas VMM provides a trajectory-conditioned velocity correction that is more closely aligned with a fixed deterministic ODE schedule. We use VMM in both warm-up and unrolling so that the distillation objective remains unchanged across the two stages.

\end{document}